\documentclass[11pt]{article}

\usepackage[preprint]{acl}

\usepackage{times}
\usepackage{latexsym}

\usepackage[T1]{fontenc}

\usepackage[utf8]{inputenc}

\usepackage{microtype}

\usepackage{inconsolata}

\usepackage{graphicx}
\usepackage{hyperref}
\usepackage{url}

\usepackage{graphicx}
\usepackage[utf8]{inputenc} 
\usepackage[T1]{fontenc}    
\usepackage{hyperref}       
\usepackage{url}            
\usepackage{booktabs}       
\usepackage{amsfonts}       
\usepackage{nicefrac}       
\usepackage{microtype}      
\usepackage{xcolor}         
\usepackage{amssymb}
\usepackage{amsmath}
\usepackage{multirow}
\usepackage{makecell}
\usepackage{wrapfig}
\usepackage{subcaption}
\usepackage{bm}
\usepackage{bbding}
\usepackage{colortbl}
\usepackage{amsfonts,amssymb}
\usepackage[most]{tcolorbox}
\usepackage{mdframed}
\usepackage{enumitem}
\usepackage{array}       
\usepackage{caption}
\usepackage{multicol}
\usepackage{arydshln} 
\usepackage{longtable}
\usepackage{multirow, tabularx, array, xcolor, colortbl}

\definecolor{modelblue}{RGB}{217,237,247} 

\definecolor{lightgrey}{gray}{0.965}
\definecolor{Gray}{gray}{0.90}
\definecolor{lightgrey}{gray}{0.965}
\definecolor{middlegrey}{gray}{0.95}

\usepackage{pifont}                
\newcommand{\cmark}{\ding{51}}      
\newcommand{\xmark}{\ding{55}}

\newcommand{\modelname}{MSEarth}
\newcommand{\qatype}{reasoning}

\title{\modelname{}: A Multimodal Benchmark for Earth Science Phenomenon Discovery with MLLMs}

\author{Xiangyu Zhao\textsuperscript{1,2}\footnotemark[2], Wanghan Xu\textsuperscript{2,3}\footnotemark[2], Bo LIU\textsuperscript{1}, Yuhao Zhou\textsuperscript{2},  {Fenghua Ling\textsuperscript{2}}, \\ \textbf{Ben Fei\textsuperscript{2, 4}}, \textbf{Xiaoyu Yue\textsuperscript{2}}, \textbf{Lei Bai\textsuperscript{2}}, \textbf{Wenlong Zhang\textsuperscript{2} \Envelope},  \textbf{Xiao-Ming Wu\textsuperscript{1}}
\\
\textsuperscript{1}The Hong Kong Polytechnic University \quad \textsuperscript{2}Shanghai Artificial Intelligence Laboratory \\ \textsuperscript{3}Shanghai Jiao Tong University \quad \textsuperscript{4}The Chinese University of Hong Kong
\\
\texttt{zhangwenlong@pjlab.org.cn, xiang-yu.zhao@connect.polyu.hk} \\
}

\begin{document}

\maketitle

\renewcommand{\thefootnote}{\fnsymbol{footnote}}
\footnotetext[2]{Equal Contribution.}

\begin{abstract}

The rapid advancement of multimodal large language models (MLLMs) offers new opportunities for complex scientific challenges, yet their application in earth science—especially at the graduate level—remains underexplored due to a lack of benchmarks reflecting the depth and complexity of geoscientific reasoning. Existing datasets often rely on synthetic data or simple figure-caption pairs, failing to capture the nuanced reasoning required for real-world applications. To address this, we introduce \modelname{}, a multimodal scientific dataset and benchmark curated from high-quality, open-access publications. Covering the five major spheres of Earth science—atmosphere, cryosphere, hydrosphere, lithosphere, and biosphere—\modelname{} features over 289K figures with refined captions enriched by contextual discussions and reasoning from the original papers. The benchmark supports tasks such as scientific figure captioning, multiple choice questions, and open-ended reasoning, providing a scalable, high-fidelity resource for developing and evaluating MLLMs in scientific reasoning. \modelname{} is publicly available to foster further research and innovation: \url{https://github.com/xiangyu-mm/MSEarth}.
\end{abstract}

\section{Introduction}

The advent of multimodal large language models (MLLMs) \citep{liu2023visual,liang2024survey} has transformed AI, enabling major advances across scientific fields. Examples include ChemVLM \citep{li2025chemvlm} in chemistry, GeoChat \citep{kuckreja2024geochat} in geography, and WeatherQA \citep{ma2024weatherqa} in atmospheric science. These models perform domain-specific visual question answering by integrating specialized knowledge: ChemVLM supports analysis of molecular structures, reactions, and chemistry exam questions, while WeatherQA enables reasoning about severe weather events in real-world settings.

Building MLLMs that understand advanced geoscientific knowledge requires rigorous datasets and benchmarks to improve performance on complex, discipline-specific problems. As shown in Table~\ref{tab:datasets}, existing benchmarks often use synthetic data or high school/undergraduate textbook materials \citep{lu2022learn,yue2024mmmu}, which lack the depth needed for professional, graduate-level tasks. Recent work \citep{li2024mmsci,roberts2024scifibench,li2024multimodal} uses academic papers to build multimodal scientific benchmarks, leveraging graduate-level complexity, but typically extracts only figures and captions and ignores key reasoning in the surrounding text. As a result, tasks are often oversimplified as basic \textit{figure-caption matching}, offering limited insight into a model’s reasoning ability.

\begin{figure*}[h]
	\centering
	\includegraphics[width=1.0\textwidth]{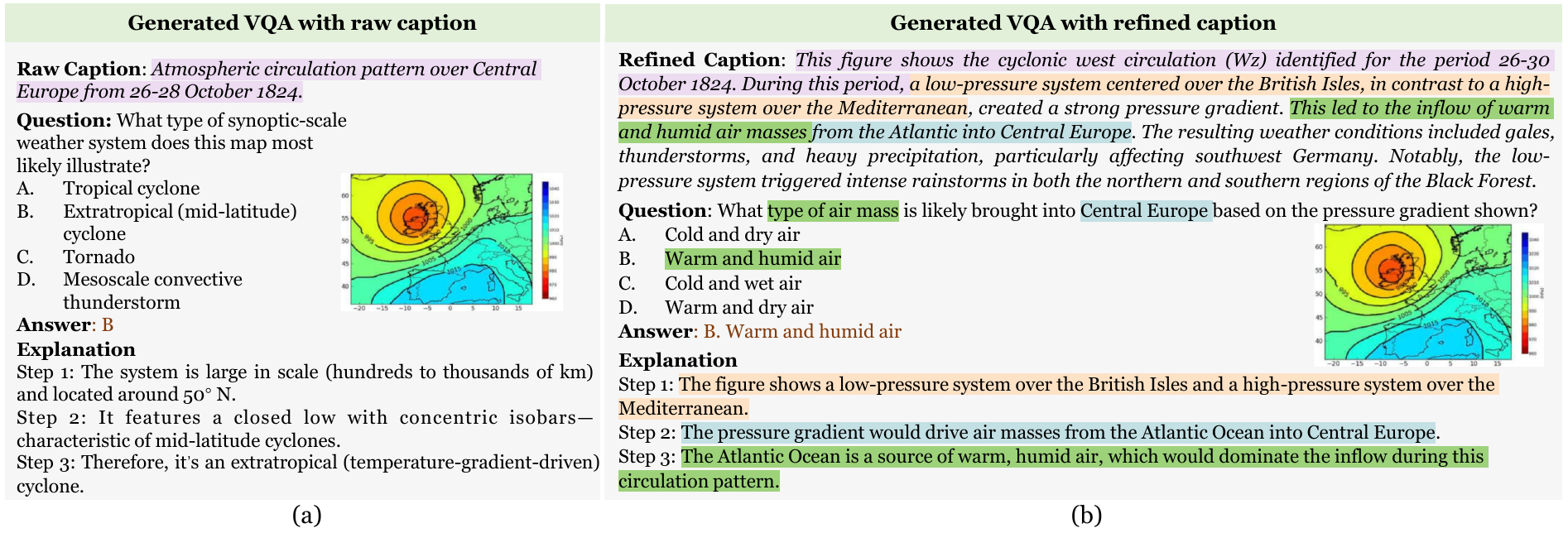}
        \caption{Illustration of VQA generation methodologies: (a) VQA relying exclusively on figure captions, and (b) VQA utilizing refined captions that integrate figure captions with content from academic papers. Highlighted areas denote questions and answers supported by evidence.
        }
        \label{intro_case}
\end{figure*}

A further challenge is designing questions that rigorously evaluate MLLMs’ data-analysis ability to uncover Earth-science phenomena from observational imagery. In scientific papers, images are often paired with hypotheses, evidence, analyses, and conclusions that are mainly in the main text rather than captions. As Figure~\ref{intro_case}(a) shows, caption-only question generation oversimplifies tasks: question quality and difficulty are constrained by generation models, and missing paper context makes verification difficult. Existing benchmarks thus neglect the high knowledge density of scientific reasoning, raising the challenge: \textit{How to effectively align high-value Earth science images with long-context information for phenomena uncovering?}

\begin{table*}[ht]
\centering
\resizebox{\textwidth}{!}{
\begin{tabular}{l c c c c c c c}
\toprule
{\bf Benchmark Dataset} & {\bf Science Topics} &  {\bf Tasks} & {\bf Difficulty} & {\bf \#Ques. } & {\bf Originality} & {\bf Validated } 
\\ \midrule

\multicolumn{7}{c}{\textbf{Multimodal Scientific benchmarks}} \\ \midrule

ScienceQA~\citep{lu2022learn} & General Science & MCQ & Primary & 21,208 & \xmark & \cmark \\

SceMQA~\citep{liang2024scemqa} & General Science & MCQ,OE & Pre-College & 1,045 & \xmark & \cmark \\

Mmmu~\citep{yue2024mmmu} & General Science & MCQ,OE & College-level & 11,550 & \xmark & \cmark \\

OlympiadBench~\citep{he2024olympiadbench} & Math, Physics & OE & Competition & 8,476 & \xmark & \cmark \\

EMMA~\citep{hao2025can} & General Science & MCQ & Mixed & 2,788 & \cmark & \cmark \\

\midrule
\multicolumn{7}{c}{\textbf{Paper-Based Multimodal Scientific benchmarks}} \\ \midrule

SciFIBench~\citep{roberts2024scifibench} & General Science & MCQ & Graduate-Level & 2,000 & \xmark & \cmark \\

ArxivCap/QA~\citep{li2024multimodal} & General Science & CG,MCQ & Graduate-Level & 32K/100K & \cmark & \xmark \\

MMSci~\citep{li2024mmsci} & General Science & CG,MCQ & Graduate-Level & 1,079,797 & \xmark & \xmark \\

\modelname{}(Ours) & Earth Science & CG,MCQ,OE & Graduate-Level & 448,980 & \cmark & \cmark \\
\bottomrule
\end{tabular}
}
\caption{
{Comparison with previous multimodal scientific benchmarks. Task types include OE (Open-ended QA), MCQ (Multiple-choice QA), and CG (Caption Generation). Originality refers to newly constructed QAs rather than those collected from existing benchmarks or textbooks. Validated indicates whether the QAs are verified by experts or supported by evidence from papers.}
}
\label{tab:datasets}
\end{table*}

We address this with a new benchmark construction approach featuring two innovations. First, we introduce the \textit{refined caption}. Observational images visualize phenomena, while deeper insights—hypotheses, supporting evidence, analytical reasoning, and conclusions—are embedded in the paper body. Raw captions are brief and lack context for complex reasoning; refined captions combine the raw caption with relevant domain-specific information extracted from the paper to provide a more complete, scientifically meaningful image description. As in Figure~\ref{intro_case}(b), refined-caption-based questions are higher quality and can be supported and validated by paper content, ensuring professionalism and accuracy. Second, we apply rigorous quality control by combining multi-agent automated evaluation with expert evaluation, ensuring questions are relevant, coherent, and matched to professional-level geoscientific phenomena uncovering. This improves reliability and accuracy beyond current LLM-based scientific question generation, producing a robust and valuable dataset for advancing MLLMs in scientific domains.

Using this adaptive annotation methodology, we present \modelname{}, a comprehensive multimodal benchmark for graduate-level Earth science. It is built from 64,560 open-access publications across five spheres, eight subjects, and 66 sub-subjects, from which we extract 289,891 figures. We augment figures with refined captions averaging 136.29 tokens (vs. 37.56 originally). The test set unifies scientific figure captioning with multiple-choice and open-ended reasoning tasks for holistic, interdisciplinary evaluation, enabling rigorous assessment of MLLMs in professional geoscientific contexts and filling a key gap in graduate-level multimodal benchmarks. \modelname{} also provides a scalable, high-fidelity pipeline for domain-specific scientific benchmark construction. Using this framework, we create a resource-rich training dataset spanning captioning, open-ended QA, and multiple-choice tasks for post-training (e.g., instruction tuning and GRPO-based reinforcement learning), yielding substantial gains across tasks for open-source models. Our contributions are:

\noindent \textbf{Development of a Scalable Adaptive Framework:} We introduce a semi-automated tool for automatic VQA generation and machine-assisted filtering, providing a robust, scalable method for high-fidelity domain-specific benchmarks extensible to other scientific fields.

\noindent \textbf{High-Quality benchmark Resources for Earth Science MLLMs:} We provide an expert-annotated graduate-level Earth science benchmark and a diverse training corpus (captioning, open-ended QA, multiple-choice) to support advanced post-training of MLLMs.

\noindent \textbf{Comprehensive Evaluation and Validation of State-of-the-Art MLLMs:} We extensively evaluate MLLMs on \modelname{}, offering insights into limitations and future directions, and demonstrate the effectiveness of our training data by building a state-of-the-art baseline model.
 
\section{Related Works}

\textbf{Multimodal Scientific Datasets and Benchmarks.}
Numerous multimodal benchmarks have been developed to evaluate scientific understanding across various domains. These benchmarks often integrate text, images, and other modalities to assess models' reasoning and cross-modal capabilities. However, their creation typically requires significant manual effort in data collection and validation. ScienceQA~\citep{lu2022learn} is an early multimodal benchmark that features multiple-choice questions (MCQs) collected from online resources and manually filtered for quality. It covers general science topics such as physics, chemistry, and biology, with a focus on elementary and high school-level reasoning. SceMQA~\citep{liang2024scemqa} and Mmmu~\citep{liang2024scemqa} extended this by incorporating both MCQs and open-ended questions (OE) from textbooks and online resources, targeting pre-college and college-level difficulty. OlympiadBench~\citep{he2024olympiadbench} introduced competition-level problems in mathematics and physics, offering open-ended tasks sourced from Olympiad exams. These problems are highly challenging but limited to specific domains. More recently, EMMA~\citep{hao2025can} combined manually designed questions with existing benchmarks, covering a broader range of topics with mixed difficulty levels. In contrast, our objective is to enhance models’ ability to comprehend multimodal, complex scientific problems—drawn from high-quality research papers in earth science—that demand graduate-level, domain-specific expertise.

\textbf{Paper-Based Multimodal Scientific Datasets and Benchmarks.}
Benchmarks based on academic papers aim to leverage the rich, domain-specific content found in scientific literature. FigureSeer~\citep{siegel2016figureseer} first extracts figures from academic papers, focusing on chart figures to evaluate the understanding of chart figures. SciFiBench~\citep{roberts2024scifibench} extended this by introducing figure-to-caption and caption-to-figure matching tasks, while MMSci~\citep{li2024mmsci} further advanced this approach using figures from Nature papers. However, these benchmarks lack original questions, limiting their ability to assess advanced reasoning and contextual understanding. ArxivQA/Cap~\citep{li2024multimodal} expanded the scope by generating new questions for figures from 32 subjects on arXiv. However, these questions were generated solely using the inherent capabilities of GPT-4V~\citep{achiam2023gpt} and did not have contextual support from the relevant text in the papers, raising concerns about their scientific validity. In contrast, our proposed benchmark, \modelname{}, addresses these limitations by introducing original, evidence-supported questions grounded in refined captions. This approach enables a rigorous evaluation of MLLMs in professional-level geoscientific applications.

\begin{figure*}[t]
	\centering
	\includegraphics[width=\textwidth]{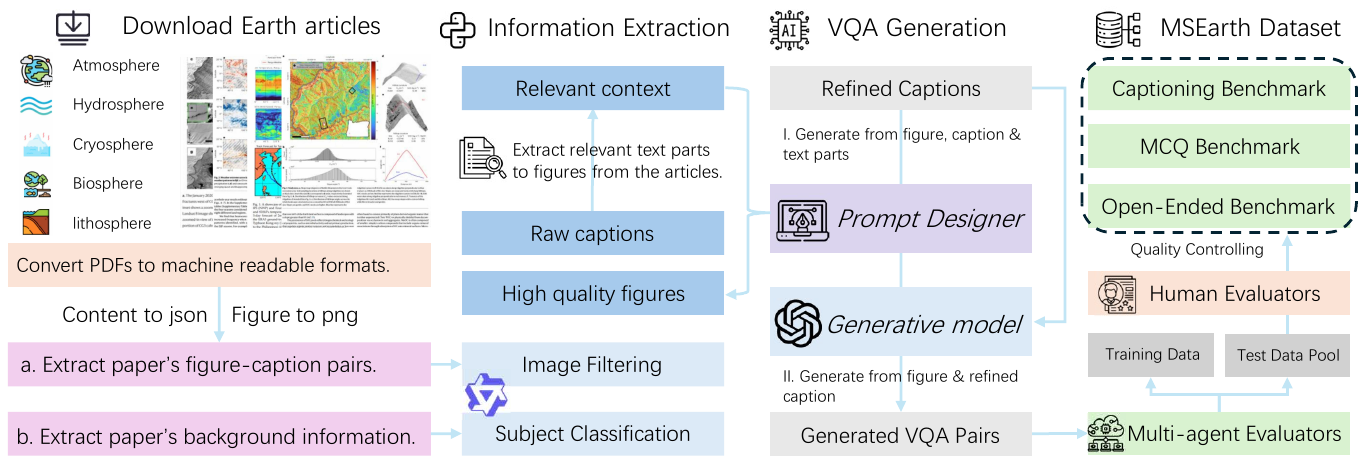}
        \caption{Data curation process for \modelname{}. The two parts on the left represent data preprocessing, while the two parts on the right encompass the automated generation of VQA and expert-AI collaborative filtering.
        }
        \vspace{-3mm}
    \label{fig: method}
 \end{figure*}

\section{MSEarth - A Multimodal Scientific Benchmark for Earth Science}
This section provides a detailed overview of the construction process for \modelname{}. As illustrated in Figure~\ref{fig: method}, we outline the framework used to develop \modelname{} from open-access scientific publications. The section is organized into three main parts: first, we detail the data collection and preprocessing steps. Next, we elaborate on the construction procedures for the two D\&B within \modelname{}, namely MSEarthCap and MSEarthQA. Finally, we describe the process of ensuring the reliability of the test data, which involves expert annotation and manual screening of the sampled test data.

\subsection{Data Preparation}
The first part of the D\&B construction focuses on data collection and preprocessing.
The data collection begins with more than 400K Earth science papers obtained in PDF format. These are uniformly converted into structured JSON text using the MinerU~\citep{wang2024mineru} parser. To classify the papers, semantic similarity is calculated between the abstracts and keywords from the five Earth spheres: hydrosphere, biosphere, lithosphere, atmosphere, and cryosphere. Based on this, the papers are assigned to respective disciplinary categories. Details are provided in Appendix~\ref{paper_filter}. We then selected papers based on the criterion of containing high-quality, Earth science-related images, resulting in a subset of around 83k papers. Specifically, Qwen-2.5-VL-72B~\citep{bai2025qwen2} is utilized to filter and select images, with the filtering prompts detailed in the Appendix~\ref{image_filter}.

\subsection{MSEarthCap}

\textbf{Figure-Caption Extraction}: Figures and their corresponding captions are extracted from the JSON files processed by MinerU. As shown in Appendix~\ref{format_concersion}, MinerU has already extracted the figures along with their original captions, which can be directly utilized for subsequent processing. To ensure accurate alignment between figures and their references within the text, we employ a regex-based method to identify the labels of each figure. This approach enables precise matching between the figures and the relevant sections of the articles.

\textbf{Relevant Context Extraction}: To further enrich the captions with contextual information, we use the figure labels obtained in the previous step to perform approximate matching against the main body of the paper. Since MinerU processes papers with segmented paragraphs, we apply regular expression matching to each paragraph to extract contextual text that references the target figure. This ensures the inclusion of descriptions and reasoning associated with each figure within the paper. To guarantee that the extracted context provides sufficient detail about the target figure, only paragraphs exceeding two sentences were included in the final dataset. From this filtered subset, we selected around 64K papers that met the criteria for subsequent processing. For more details, refer to Appendix~\ref{content_filter}.

\textbf{Refined Caption Generation}: To create professional-level figure descriptions, we employ GPT-4o for refined caption generation. The model takes as input the extracted figure, its original caption, and the contextual text from the relevant sections of the paper. Figure refinement is performed only for data that includes valid relevant contexts. The specific prompts used for this process are detailed in the Appendix~\ref{prompt_design}. After statistical analysis, we observe that the average word length of the raw captions is 37.56, while the average length of the refined captions increases to 136.29, reflecting the incorporation of richer, domain-specific content.


\subsection{MSEarthQA}

To generate high-quality multiple-choice questions (MCQs) and open-ended questions, we use a question generation pipeline that takes the figure, its original caption, and the refined caption as input. The generation prompts, detailed in the Appendix~\ref{prompt_design}, are crafted to encourage the model to highlight differences between the original and refined captions, ensuring that the generated questions are grounded in evidence from the paper. The questions are constructed using GPT-4o to maintain high detail and relevance. However, due to inherent challenges such as self-inconsistency and uncertainty in the generation process, the generated questions undergo an automated and expert validation process to ensure quality.

\begin{figure*}[t]
	\centering
	\includegraphics[width=\textwidth]{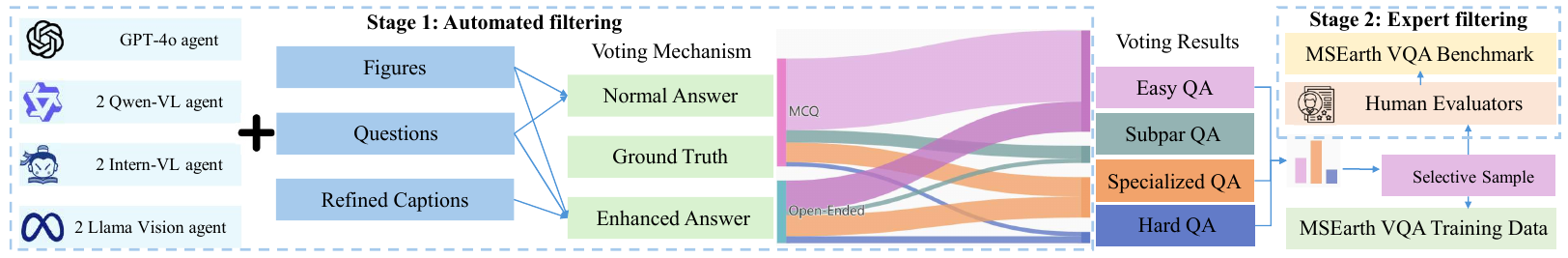}
        \caption{Overall approach of our multi-agent, voting-based approach to automate the validation of generated questions.
        }
    \label{fig: voting}
 \end{figure*}

\subsubsection{Automated Validation}  
Inspired by the LLM Voting~\citep{yang2024llm,lee2025reliable,kaesberg2025voting} method, we developed a multi-agent, voting-based approach to automate the validation of generated questions. Specifically, we employ a \textit{Majority Voting} strategy, where multiple agents independently generate responses, and the final decision is based on the majority consensus of these agents.  In our setup, we utilize the following MLLMs for decision-making: Qwen2.5-VL-72B, Qwen2.5-VL-7B, InternVL2.5-7B, InternVL2.5-78B, and GPT-4o.
A key aspect of our evaluation process is the use of the \textit{refined caption}, which incorporates scientists' reasoning and insights about the figure extracted from the paper. This refined caption provides additional context and domain-specific information that goes beyond the original caption. By comparing model performance with and without the refined caption, we can assess the quality of the questions and determine whether they test the model's ability to grasp scientific reasoning and insights. The detailed decision-making process is outlined below:

\noindent \textbf{Phase A:}  
The question and original caption are provided to a suite of models $\{M_{1}, M_{2}, ..., M_{n}\}$. The correctness of the model responses is used to evaluate the types and quality of the questions. A threshold of 60\% is defined for supermajority voting. Specifically, if more than 40\% of the models produce incorrect responses, the question is flagged for further analysis. Additionally, we discard questions that all models answer correctly, as these questions do not contribute to the effective testing or training of the models. Questions identified through this process are categorized as either potentially difficult or of poor quality, with the distinction made in subsequent phases.

\noindent \textbf{Phase B:}  
In this phase, the question and refined caption are provided to the same suite of models. If more than 60\% of the models answer the question correctly with the refined caption, it indicates that the question requires relatively specialized scientific knowledge to answer. Such questions are categorized as \textit{specialized QA}, as their answers rely on the model's ability to understand and apply specific domain knowledge rather than simply perceiving the image or relying on commonsense reasoning. Questions that fail this phase proceed to the next stage for further evaluation.

\noindent \textbf{Phase C:}  
In this phase, only models with 70B+ parameters are used for voting, including GPT-4o (the same model used for question generation), InternVL2.5-78B, and Qwen2.5-VL-72B. The question and its refined caption are provided to these large-scale models. If more than 60\% of the large models answer the question correctly, it suggests that the difficulty of the question likely lies in the model's ability to perceive and interpret the image content. Such questions are categorized as \textit{hard QA}. Subsequent human validation will involve sampling and additional annotation across QA filtered in all phases to ensure the overall quality and accuracy of the benchmark.


This pipeline identifies high-quality questions by filtering out overly simplistic or poorly constructed ones.  
In \textbf{Phase A}, approximately 70\% of the questions were categorized as \textit{easy}, as most models could answer them correctly without refined captions.  
After \textbf{Phase B}, around 20\% were classified as \textit{specialized QA}, where refined captions enabled correct answers, indicating the need for domain-specific knowledge.  
In \textbf{Phase C}, 5\% were labeled as \textit{hard QA}, requiring high-performing models to interpret image content accurately, suggesting that these questions test the model's ability to perceive and interpret image content.  
The remaining 5\% were deemed flawed and discarded. Detailed processes and examples are provided in Appendix~\ref{agnet-voting}. For the training dataset, we sampled more than 150K VQA pairs, including both multiple-choice and open-ended questions. Of these, 20\% were drawn from Phase A and 80\% from Phase B, ensuring a balanced distribution of question difficulty.

\begin{table}[htb]
    \centering
    \fontsize{8.0pt}{\baselineskip}\selectfont
    \setlength{\tabcolsep}{4pt} 
    \renewcommand\arraystretch{1.1}
    \begin{tabular}{l cc}
        \toprule
        \textbf{Statistic} & \textbf{Training Set} & \textbf{Test Set} \\
        \midrule
        Captioning as QA & 289,891 & 3,000 \\
        MCQ & 102,753 & 2,784 \\
        Open-Ended QA & 49,141 & 1,411 \\
        \cmidrule(lr){2-3}
        \textbf{Total Questions} & \textbf{441,785} & \textbf{7,195} \\
        \midrule
        Total Figures & \multicolumn{2}{c}{289,891} \\
        Avg. Caption Length & \multicolumn{2}{c}{37.56 (Original) / 136.29 (Refined)} \\
        \bottomrule
    \end{tabular}
    \caption{Main statistics in \modelname{}-Bench. The dataset covers 66 subjects and 64,560 articles.}
    \label{tab:all_stat}
\end{table}

\subsubsection{Expert Validation}
Ensuring that synthetic data closely mirrors real-world distributions is critical for evaluation tasks. To achieve this, domain experts are engaged to review and annotate the curated QA pairs for accuracy and relevance. During this process, low-quality or invalid questions are identified and filtered out to ensure the overall quality of the dataset. The annotation process is conducted from two perspectives: image types and question types. For image types, we categorize the data into three groups: single-image question answering, single-image-focused question answering within multi-image figures, and multi-image relational question answering. For question types, we define two categories: scientific \qatype{} and perception questions. Scientific reasoning questions are constructed based on inferences or scientific discoveries presented in research papers, making them more specialized and challenging. In contrast, image perception questions focus on interpreting images and require less background knowledge of scientific concepts. This expert-AI collaborative process, combined with rigorous quality control, results in a high-quality dataset comprising 1,500 open-ended questions and 3,000 MCQs, forming the MSEarthQA benchmark. The annotated results are summarized in Table~\ref{tab:all_stat}, with further details and analysis provided in Appendix~\ref{expert_val} and Table~\ref{tab:statistics}.

\begin{figure*}[t] 
 \centering
 
 \begin{minipage}{0.48\textwidth} 
 \centering

 \vspace{2mm} 
 \fontsize{8.0pt}{10.0pt}\selectfont 
 \renewcommand\tabcolsep{3.0pt} 
 \renewcommand\arraystretch{1.0} 
 \begin{tabular}{lc}
 \toprule
 \textbf{Statistic} & \textbf{Number} \\
 \midrule
 Total questions & 7,195 \\
 \midrule
 MCQ & 2,784 \\
 Questions with single images & 1,255 (45.1\%) \\
 Questions with multiple images & 1,529 (54.9\%) \\
 ~~~~~* Single-image focus & $\approx$1,164 (41.8\%) \\
 ~~~~~* Multi-image relational & $\approx$365 (13.1\%) \\
 \hdashline
 Discovery Question & 2,117 (76.0\%) \\
 Perception Question & 667 (24.0\%) \\
 \midrule
 Open-Ended & 1,411 \\
 Questions with single images & 679 (45.1\%) \\
 Questions with multiple images & 832 (45.1\%) \\
 ~~~~~* Single-image focus & $\approx$619 (41.8\%) \\
 ~~~~~* Multi-image relational & $\approx$113 (13.1\%) \\
 Captioning & 3,000 \\
 \midrule
 Average caption length & 37.71 \\
 Average refined caption length & 137.47 \\
 \bottomrule
 \end{tabular}
 \captionof{table}{Main statistics in \modelname{}.}
 \label{tab:statistics}
 \end{minipage} 
 \hfill 
 \begin{minipage}{0.48\textwidth} 
 \centering
 \includegraphics[width=0.95\linewidth]{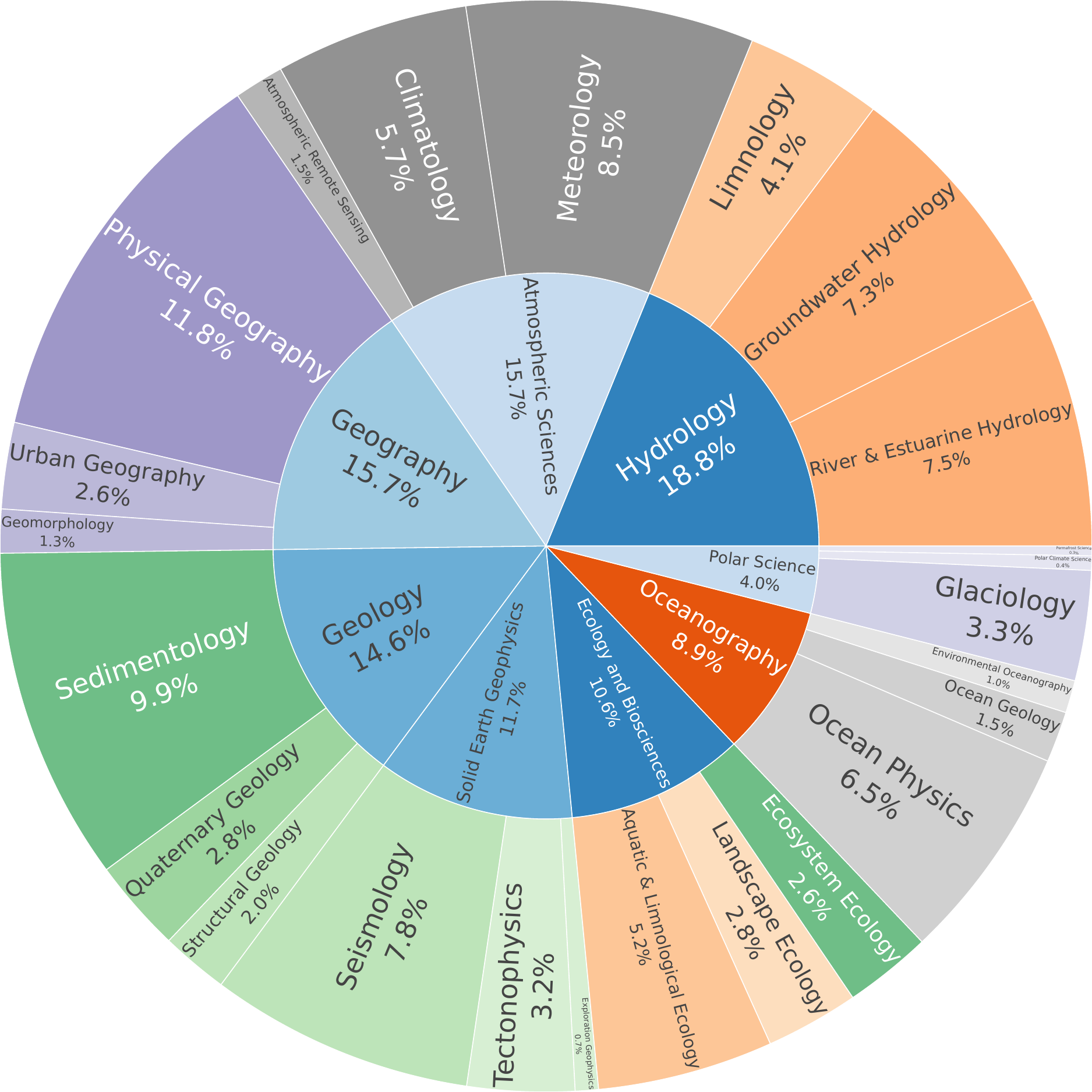}
 \caption{Subjects distribution in \modelname{}.}
 \label{fig:que_pie}
 \end{minipage}
 
\end{figure*}

\section{Experiments}


\subsection{Evaluated Models}
We evaluate different families of MLLMs on our benchmark. We evaluate the following closed-source models: GPT-4 series~\citep{hurst2024gpt}, Gemini-2.5 series~\citep{team2023gemini} and Claude-3 series~\citep{anthropic2024claude}. We also evaluate the following open-source models: LLaVA-OneVision~\citep{li2024llava}, Qwen-2.5-VL~\citep{yang2024qwen2}, InternVL2.5/3/S1~\citep{chen2024expanding,zhu2025internvl3,bai2025intern} and Llama-3.2-Vision-Instruct~\citep{grattafiori2024llama}. We use chat/instruction-tuned variants of each model and compare the performance of multiple model sizes where available. Details can be found in Appendix~\ref{baselines}. To validate the effectiveness of our training data, we conducted post-training on the Intern-s1-mini and Qwen-2.5-VL-7b models. Specifically, we employed instruct tuning method for fine-tuning on captioning and open-ended QA tasks. For the MCQ task, we applied GRPO~\citep{shao2024deepseekmath} reinforcement learning method.

\subsection{ Evaluation Metrics}

Both captioning and open-ended QA tasks require generating freeform textual outputs grounded in complex scientific data and reasoning. To evaluate these tasks, we use lexical overlap-based metrics such as BLEU~\citep{papineni2002bleu}, ROUGE~\citep{lin2004rouge}, and METEOR~\citep{banerjee2005meteor} for surface-level similarity, while BERTScore~\citep{zhang2019bertscore} assesses deeper semantic alignment. Additionally, following G-Eval~\citep{liu2023g}, we utilize the Qwen2.5-VL-72B model with a specialized prompt to compute a factual scientific score. For the captioning task, we define a Cap-Eval score ranging from 1 to 5, where higher scores indicate better caption quality. For the open-ended QA task, we introduce OE-Eval, which evaluates the reasonableness of generated answers using a binary 0/1 scoring system.

Evaluating MCQs is relatively straightforward, as these tasks require selecting the correct answer from a predefined set of options. In our experiments, models were guided by carefully structured prompts to ensure adherence to a specific output format. Regular expression rules were employed to extract the selected choice, ensuring strict alignment with the predefined answer format. Detailed evaluation metrics are provided in Appendix~\ref{metrics}.



\begin{table*}[ht]
\scriptsize
\centering
\resizebox{\textwidth}{!}{
\begin{tabular}{lcccccc}
\toprule

\multirow{2}{*}{\textbf{Model}} & \multicolumn{3}{c}{\textbf{Image-Type}} & \multicolumn{2}{c}{\textbf{Task Type}} & \multicolumn{1}{c}{\textbf{Overall}} \\

 & \textbf{\textsc{Single}} & \textbf{\textsc{Multi}} & \textbf{\textsc{Cross}}  & \textbf{\textsc{\qatype{}}} & \textbf{\textsc{Percept}} & \textbf{\textsc{ACC}} \\
\midrule

\rowcolor{Gray} \multicolumn{7}{c}{\textit{\textbf{Open-source Models}}} \\

LLaVA-onvision-72B  & 53.55 & 49.48 & 47.95 & 46.58 & 65.52 & 51.11 \\
Qwen2.5-VL-7B  & 47.65 & 44.07 & 37.53 & 40.53 & 58.47 & 44.83 \\
Qwen2.5-VL-72B  & 52.11 & 50.43 & 46.30 & 44.40 & 70.46 & 50.65 \\
InternVL2-8B  & 44.86 & 43.99 & 38.36 & 38.97 & 58.47 & 43.64 \\
InternVL2.5-78B  & 53.23 & 49.74 & 44.38 & 43.17 & 74.21 & 50.61 \\
InternVL3-78B & 57.53 & 51.37 & 45.48 & 47.00 & 73.61 & 53.38 \\
Llama3.2-90B-Vision & 45.98 & 40.46 & 38.90 & 38.26 & 56.97 & 42.74 \\
DeepSeek-VL2 & 52.43 & 49.23 & 44.66 & 46.06 & 62.82 & 50.07 \\
Qwen3-VL-32B & 54.58 & 50.77 & 47.95 & 45.58 & 72.86 & 52.12 \\
GLM-4.1V-Thinking & 54.18 & 49.83 & 38.36 & 46.29 & 62.97 & 50.29 \\
Intern-S1-mini & 60.00 & 57.22 & 49.04 & 51.68 & 75.56 & 58.12 \\
Intern-S1 & 67.01 & 65.62 & 64.11 & 61.22 & 79.61 & 65.63 \\

\hdashline
Qwen-7B-MSEarth  & 57.61 & 52.75 & 45.20 & 50.68 & 64.32 & 53.95 \\
Intern-S1-mini-MSEarth  & \underline{65.49} & \underline{62.97} & \underline{58.63} & \underline{58.81} & {78.56} & \underline{63.54} \\
\midrule
\rowcolor{Gray}
\multicolumn{7}{c}{\textit{\textbf{Proprietary Models}}} \\

Gemini-2.5-Flash & 58.33 & 54.55 & {53.42} & 49.98 & 75.56 & 56.11 \\
Gemini-2.5-Flash-Thinking & {60.64} & 54.64 & 53.70 & {51.35} & 75.86 & 57.22 \\
Gemini-2.5-Pro-Thinking & {64.78} & {59.36} & {55.34} & {56.31} & 77.06 & {61.28} \\
Claude-3.5-Haiku & 49.48 & 47.16 & 42.47 & 42.18 & 64.77 & 47.59 \\
Claude-3.7-Sonnet  & 59.52 & {56.53} & {57.53} & {51.68} & {78.11} & {58.01} \\
GPT-4o & {63.03} & 55.76 & 47.67 & 50.45 & \textbf{81.86} & 57.97 \\
GPT-5.2 & {61.99} & 59.54 & 54.52 & 54.98 & {75.86} & 59.99 \\
Gemini-3-Flash & \textbf{70.44} & \textbf{67.70} & \textbf{66.85} & \textbf{65.14} & \underline{80.51} & \textbf{68.82} \\
\bottomrule
\end{tabular}%
}
\caption{{Accuracies (\%) of different models on multiple-choice questions.} The best results are highlighted in bold, with the second-best underlined.}
\label{tab:vqa}
\end{table*}

\subsection{Main Results}


\textbf{Current models struggle with scientific question-answering tasks, particularly on questions requiring specialized knowledge and reasoning across multiple images.} The performance of MCQs is summarized in Table~\ref{tab:vqa}. The results reveal that most models do not perform exceptionally well on scientific question-answering tasks, with proprietary models generally achieving better results. Further analysis of the models' failure rates on \qatype{} and perception-based questions is provided in the Appendix (Figure~\ref{vqa_accuracy}). This analysis shows that models are more prone to errors on questions requiring specialized knowledge, underscoring significant room for improvement in scientific reasoning question-answering. In contrast, for relatively simpler perception-based questions, which require less domain-specific knowledge, the models tend to perform better. Similarly, when analyzing performance across different image types, we observe that most models achieve their best results on tasks involving single-image inputs. However, for tasks requiring multi-image inputs, particularly those that demand reasoning across multiple images to derive an answer, the models perform the worst. Additional experimental results can be found in Appendix~\ref{detailed_MCQ}.

\textbf{Proprietary models consistently outperform open-source models in both Scientific Figure Captioning and Open-Ended VQA tasks, with LLM-based metrics providing a more reasonable evaluation.} The captioning results are presented in Table~\ref{tab:combined_results}, where overlap-based, similarity-based, and LLM-based metrics exhibit similar trends, with no significant differences observed among the overlap-based and similarity-based metrics. The Gemini-3-Flash model achieves the best performance across most metrics. The LLM-based metric, designed to evaluate the professionalism and accuracy of generated captions, demonstrates greater variance compared to similarity-based metrics, making it more suitable for assessing the Scientific Figure Captioning task. Open-source models still show a noticeable gap compared to proprietary models, consistent with the findings from the MCQ results, suggesting a close interconnection between a model's understanding and reasoning capabilities. Similarly, the results for open-ended question answering, presented in Table~\ref{tab:combined_results}, show that overlap-based and similarity-based metrics tend to yield higher scores due to the shorter nature of both ground truth answers and model-generated responses. However, for open-ended questions, the focus should be on the rationality and correctness of the answers, making the LLM-based metric a more reasonable evaluation method. This metric also reveals trends consistent with the previous tasks, further highlighting the performance gap between open-source and proprietary models.

\begin{table*}[ht]
\centering
\resizebox{\textwidth}{!}{%
\begin{tabular}{l ccccc ccccc}
\toprule
\multirow{3}{*}{\textbf{Model}} & \multicolumn{5}{c}{\textbf{Open-ended QA}} & \multicolumn{5}{c}{\textbf{Figure Captioning}} \\
\cmidrule(lr){2-6} \cmidrule(lr){7-11}
 & \multicolumn{3}{c}{{Overlap}} & {Sim.} & {MLLM} & \multicolumn{3}{c}{{Overlap}} & {Sim.} & {MLLM} \\
\cmidrule(lr){2-4} \cmidrule(lr){5-5} \cmidrule(lr){6-6} \cmidrule(lr){7-9} \cmidrule(lr){10-10} \cmidrule(lr){11-11}
 & \textsc{R-L} & \textsc{Met} & \textsc{BLEU} & \textsc{BS} & \textsc{OE}(\%) & \textsc{R-L} & \textsc{Met} & \textsc{BLEU} & \textsc{BS} & \textsc{Cap} \\
\midrule

\rowcolor{Gray}
\multicolumn{11}{c}{\textit{\textbf{Open-source Models}}} \\

LLaVA-onevision-72B & 37.91 & 27.88 & 1.94 & 89.72 & 41.56 & \textbf{17.82} & 18.35 & 2.15 & 83.47 & 2.07 \\
Qwen2.5-VL-7B       & 35.24 & 29.00 & {2.40} & 88.62 & 40.68 & 15.72 & 16.57 & 1.55 & 83.82 & 2.22 \\
Qwen2.5-VL-72B      & 38.34 & 30.65 & 2.20 & 89.22 & 44.82 & 16.53 & 21.35 & 2.36 & 83.87 & 2.56 \\
InternVL2.5-8B      & 35.95 & 28.25 & 2.04 & 89.14 & 39.05 & 16.71 & 19.28 & 1.58 & 83.25 & 1.91 \\
InternVL2.5-78B     & 40.34 & 31.15 & 2.23 & {90.05} & 45.64 & 17.24 & 20.66 & 2.27 & 83.56 & 2.30 \\
InternVL3-78B       & {40.42} & 31.77 & 2.37 & {89.98} & 47.00 & 16.95 & 20.95 & 2.32 & 83.72 & 2.43 \\
Llama3.2-90B-Vision & 37.53 & 29.08 & 1.99 & 89.16 & 42.72 & 12.98 & 21.21 & 1.49 & 78.89 & 1.82 \\
DeepSeek-VL2        & 36.37 & 27.24 & 1.83 & 89.38 & 40.68 & 16.71 & 18.22 & 1.69 & 83.64 & 2.22 \\
Intern-S1-mini      & 37.46 & 29.72 & {2.34} & 89.25 & {43.69} & 17.42 & 24.18 & 2.26 & 83.65 & 2.65 \\

\hdashline
Qwen-7B-MSEarth          & 38.47 & 30.10 & 2.37 & 89.82 & 48.19 & 17.74 & 26.59 & 2.86 & 83.71 & 2.73 \\
Intern-S1-mini-MSEarth   & \underline{41.36} & 31.87 & {2.44} & \bf 90.42 & {49.74} & \underline{17.94} & \underline{27.61} & {3.25} & 83.86 & 3.02 \\

\midrule
\rowcolor{Gray}
\multicolumn{11}{c}{\textit{\textbf{Proprietary Models}}} \\

Gemini-2.5-Flash          & 40.02 & \underline{32.34} & 2.06 & 89.39 & \underline{52.00} & {17.79} & 23.01 & {3.10} & \textbf{83.96} & 2.98 \\
Gemini-2.5-Flash-Thinking & 39.47 & 32.41 & 2.52 & 90.11 & 52.40 & 17.42 & 23.47 & 2.97 & 83.85 & {3.04} \\
Gemini-2.5-Pro-Thinking   & 40.39 & 31.73 & \underline{2.50} & 88.93 & 47.70 & 17.15 & {27.45} & \underline{3.33} & 83.62 & \underline{3.35} \\


Claude-3.7-Sonnet         & 40.21 & 30.75 & 1.73 & 89.37 & {48.33} & 15.94 & 21.15 & 2.15 & 83.62 & 2.71 \\
GPT-4o                    & {41.03} & {32.70} & 2.04 & 89.78 & {48.55} & 16.33 & 20.19 & 2.15 & \underline{83.93} & 2.72 \\
Gemini-3-Flash       & \textbf{41.48} & \textbf{32.85} & \textbf{2.46} & \underline{90.26} & \textbf{52.61} & \textbf{17.96} & \textbf{28.32} & \textbf{3.35} & {83.92} & \textbf{3.40} \\

\bottomrule
\end{tabular}%
}
\caption{
\textbf{MLLM's performance on Scientific Open-ended QA and Figure Captioning.}
We report \textsc{RougeL} (R-L), \textsc{Meteor} (Met), \textsc{BLEU}, \textsc{BertScore} (BS), and the respective LLM-based evaluations (\textsc{OE-Eval} and \textsc{Cap-Eval}).
The best results are highlighted in bold, with the second-best underlined.
}
\label{tab:combined_results}
\end{table*}


\subsubsection{Analysis and Discussion}

As shown in Table~\ref{tab:vqa}, most MLLMs exhibit a pronounced gap between perception‐based and reasoning‐based performance. On simple visual questions—where answers depend on direct feature extraction—these models routinely exceed 75\% accuracy.  However, on scientific reasoning tasks that demand domain‐specific knowledge, their scores drop sharply (e.g.\ Gemini-2.5-Pro-Thinking: 77.06\% perception vs.\ 56.31\% reasoning).  This divergence suggests that while robust perception is a necessary foundation, it is not sufficient for Earth‐science inference: once a model surpasses the $\approx75\%$ perception threshold, further gains hinge on integrating specialized knowledge and enabling multi‐step reasoning.
By contrast, models pretrained or fine‐tuned on scientific datasets, such as Intern-S1, demonstrate substantially higher accuracy in both perception and reasoning, thereby confirming that general MLLMs lack the requisite Earth‐science expertise. Furthermore, we find that further training on the MSEarth training set boosts performance across the board, with the largest relative improvement appearing in reasoning tasks. Taken together, these results underscore the critical role of domain‐focused data and architectures in closing the reasoning gap.

Key factors driving low reasoning‐task performance include (1) insufficient coverage of Earth science content in general pretraining corpora, (2) the absence of iterative chain‐of‐thought reasoning modules in standard multimodal fusion backbones, and (3) the scarcity of annotated, multimodal datasets that provide step‐by‐step scientific rationales.

\begin{figure}[t]
    \centering
    \includegraphics[width=0.8\linewidth]{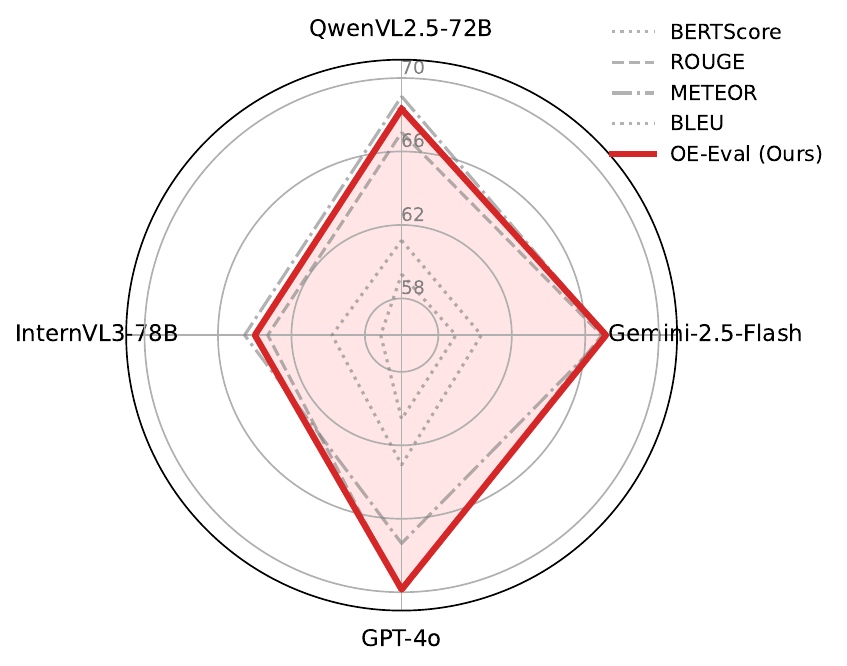}
    \caption{Comparison of Spearman correlations across different models. \textbf{OE-Eval} (red line) demonstrates consistently high correlation with human judgments compared to traditional metrics.}
    \label{fig:radar_oe_eval}
\end{figure}

\begin{table*}[h]
\centering
\small
\resizebox{\textwidth}{!}{
\begin{tabular}{lccccccccc}
\toprule
Model & Atmospheric & Solid Earth Geophysics & Geography & Ecology & Geology & Hydrology & Oceanography & Polar & All \\
\midrule
InternVL3-78B & 50.70\% & 29.73\% & 28.57\% & 47.06\% & 25.00\% & 51.02\% & 25.00\% & 30.00\% & 47.33\% \\
Gemini-2.5-Pro & 46.48\% & 58.11\% & 35.00\% & 35.71\% & 47.06\% & 50.00\% & 59.18\% & 50.00\% & 51.33\% \\
o4-mini & 50.00\% & 45.00\% & 55.88\% & 71.43\% & 49.30\% & 48.98\% & 43.33\% & 62.50\% & 53.00\% \\
Expert & 86.49\% & 85.00\% & 85.29\% & 92.86\% & 87.32\% & 85.71\% & 86.67\% & 87.50\% & 87.00\% \\
\bottomrule
\end{tabular}
}
\caption{Accuracy on MSEarth-Bench-mini across Earth science domains. Human expert scores are averages of three Ph.D.-level Earth science evaluators.}
\label{tab:msearth_mini_human}
\end{table*}

\subsubsection{MLLM-BASED METRICS}

Following LAVE~\citep{manas2024improving}, in order to assess the validity of OE-Eval, we calculated its correlation with human judgment using Spearman’s rank correlation coefficients. To validate alignment with human judgment, four Earth Science Ph.D. candidates evaluated 160 random samples from the MSEarth Open Ended benchmark across four MLLMs, as detailed in Appendix~\ref{mllm_metrics}. With an inter-annotator agreement (Krippendorff’s $\alpha$) of 0.695, the results in Figure~\ref{fig:radar_oe_eval} demonstrate that OE-Eval achieves higher consistency with human judgment than all baselines, effectively capturing the scientific nuance required by our benchmark.

\section{Human Performance Baseline}

To clarify benchmark difficulty and further justify its educational relevance, we additionally report human expert scores on the MSEarth-Bench-mini set. MSEarth-Bench-mini is constructed by sampling 300 specialized questions from the original MCQ dataset, enabling us to examine whether existing methods can improve model performance on these domain-focused problems. We hired three Ph.D. students with backgrounds in Earth sciences to evaluate the tasks, and report their average scores as a human-performance baseline in Table~\ref{tab:msearth_mini_human}. The results show that human experts consistently outperform current MLLMs across all Earth science domains.

\section{Conclusion}

We introduce \modelname{}, a graduate-level multimodal dataset designed for MLLMs in geoscientific applications. \modelname{} not only serves as a robust test set but also includes rich training resources aimed at enhancing the geoscientific understanding and reasoning capabilities of existing MLLMs. Our evaluation reveals significant gaps in current MLLMs’ ability to handle complex, graduate-level geoscientific reasoning, highlighting opportunities for improvement.  We believe \modelname{} will serve as a valuable resource for advancing MLLMs in scientific reasoning.

\section*{Acknowledgments}
The authors thank the anonymous reviewers for their insightful and constructive feedback. This work was partially supported by the Shanghai Artificial Intelligence Laboratory, the Hong Kong General Research Fund (15204225), and Project P0056021 (Parent Project P0050643) of the Otto Poon Charitable Foundation Smart Cities Research Institute, The Hong Kong Polytechnic University. This work was done during the author's internship at Shanghai Artificial Intelligence Laboratory. 

\bibliography{custom}
\newpage
\appendix



\section{Limitations}

The limitation of this work lies in the vastness and diversity of geosciences as a discipline. While we have made efforts to cover a wide range of topics, it is inevitable that certain niche or highly specialized subfields may not be adequately represented in \modelname{}.

\section{Potential Risks}
All papers used were obtained from OpenDataLab~\citep{he2024opendatalab} under the CC BY 4.0 license, which permits adaptation and redistribution with attribution. We strictly adhered to all licensing terms and usage requirements specified by OpenDataLab.
We have not identified any potential risks or negative societal impacts associated with this work. The dataset is constructed from publicly available scientific literature and is intended solely to advance research in multimodal scientific reasoning.

\section{Usage of Language Models}

We utilized a large language model (LLM) to aid in the preparation of this manuscript. Its use was limited to editorial tasks, including proofreading for typographical errors, correcting grammar, and improving the clarity and readability of the text.

\begin{figure*}[!t]
	\centering
	\includegraphics[width=1.0\textwidth]{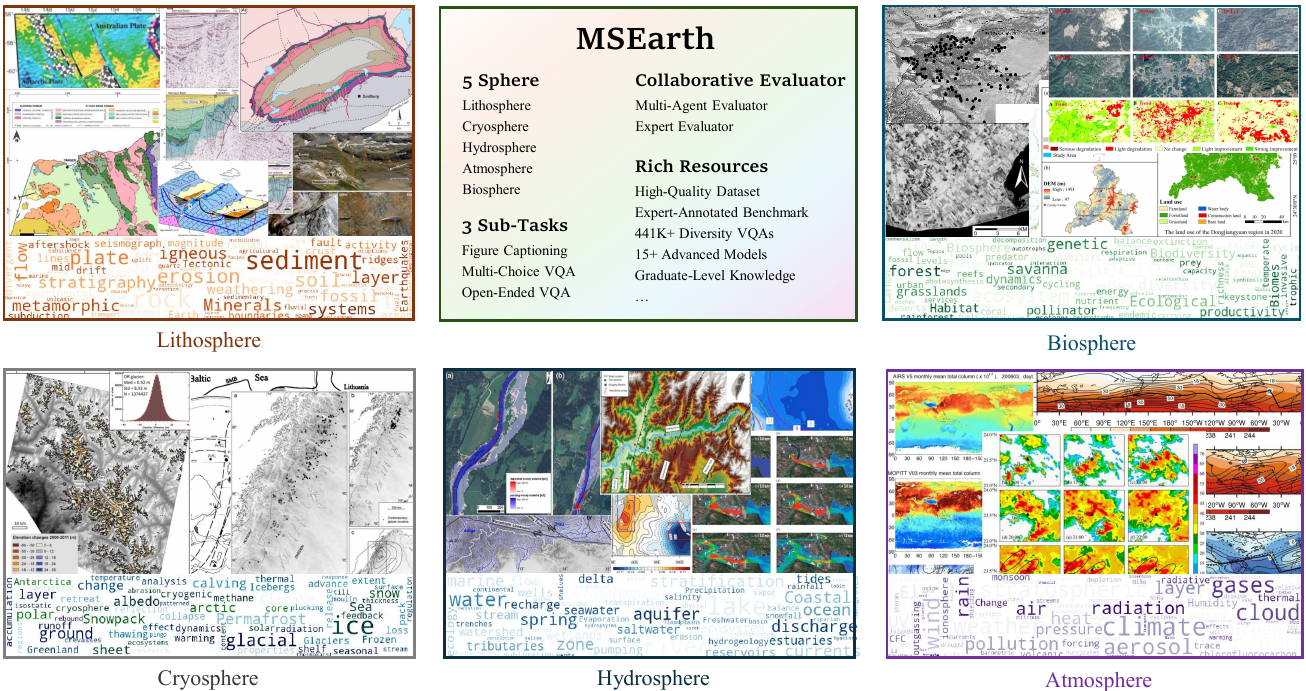}
        \caption{Illustrative examples of the diverse types of scientific figures in \modelname{}, sourced from open-access articles available from website.}
        \label{task—example}
\end{figure*}

\begin{figure*}[!t]
	\centering
	\includegraphics[width=1.0\textwidth]{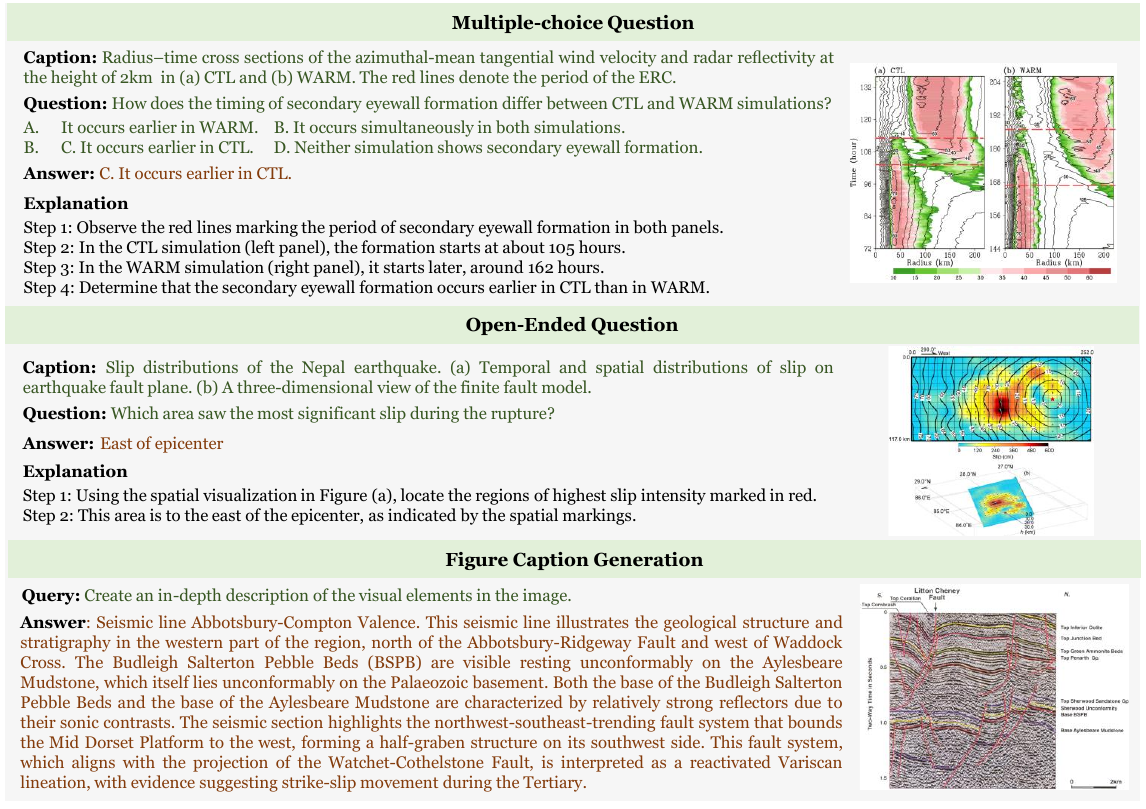}
        \caption{Examples of the three types of scientific question-answering tasks presented in our benchmark.}
        \label{task_case}
\end{figure*}

\section{Reproducibility}
All papers used were obtained from OpenDataLab~\citep{he2024opendatalab} under the CC BY 4.0 license, which permits adaptation and redistribution with attribution. We strictly adhered to all licensing terms and usage requirements specified by OpenDataLab. This work establishes a benchmark for evaluating the multimodal Earth scientific exploration capabilities of MLLMs in the field of Earth sciences. It has broader positive impacts, including promoting the responsible use of AI in scientific research and enhancing public understanding of Earth sciences. We believe \modelname{} will serve as a valuable resource for advancing multimodal language models (MLLMs) in scientific reasoning, and we plan to expand its scope to other scientific domains in future work. The benchmark is publicly available to foster further research and innovation in this field. All data in \modelname{} are released anonymously, including the complete dataset on HuggingFace (\url{https://huggingface.co/MSEarth-Data}) and all data‐processing, training, and evaluation code on Anonymous Github (\url{https://anonymous.4open.science/r/MSEarth-2B3F})

\section{Benchmark Details}
\label{benchmark}













\subsection{Field Explanation of \modelname{}}

In Table~\ref{tab:task_field_descriptions} and Figure~\ref{task_case}, we provide an explanation of each field for the three tasks in \modelname{}.

\begin{table*}[h]
\centering
\resizebox{\textwidth}{!}{%
\begin{tabular}{l c p{10cm}}
\toprule
\textbf{Field Name} & \multicolumn{1}{c}{\textbf{Input}} & \multicolumn{1}{c}{\textbf{Description}} \\
\midrule

\rowcolor{Gray}
\multicolumn{3}{c}{\textit{\textbf{Multiple-choice Question}}} \\

\texttt{question\_id} & \xmark & The unique identifier for the question. \\

\texttt{query} & \cmark & Contains the original caption, question, and options. \\

\texttt{response} & \xmark & The correct answer to the question. \\

\texttt{images} & \cmark & The file path to the associated image(s). \\

\texttt{refined\_caption} & \xmark & The enhanced image description based on the paper content. \\

\texttt{classification} & \xmark & The classification of the question, including its domain and discipline. \\

\texttt{reasoning\_chain} & \xmark & The reasoning steps to arrive at the answer. \\

\midrule
\rowcolor{Gray}
\multicolumn{3}{c}{\textit{\textbf{Open-Ended Question}}} \\

\texttt{question\_id} & \xmark & The unique identifier for the question. \\

\texttt{query} & \cmark & Contains the original caption and the question. \\

\texttt{response} & \xmark & The correct answer to the question. \\

\texttt{images} & \cmark & The file path to the associated image(s). \\

\texttt{refined\_caption} & \xmark & The enhanced image description based on the paper content. \\

\texttt{classification} & \xmark & The classification of the question, including its domain and discipline. \\

\texttt{reasoning\_chain} & \xmark & The reasoning steps to arrive at the answer. \\

\midrule
\rowcolor{Gray}
\multicolumn{3}{c}{\textit{\textbf{Caption Generation}}} \\

\texttt{question\_id} & \xmark & The unique identifier for the question. \\

\texttt{query} & \cmark & The question. \\

\texttt{response} & \xmark & The correct answer (refined caption). \\

\texttt{images} & \cmark & The file path to the associated image(s). \\

\texttt{context} & \xmark & The text from the paper that describes the image. \\

\texttt{original\_caption} & \xmark & The original caption of the image. \\

\texttt{classification} & \xmark & The classification of the question, including its domain and discipline. \\

\bottomrule
\end{tabular}%
}

\caption{
\textbf{Field Descriptions for Different Tasks.} The table provides details about each field, whether it is used as input, and its description. Fields are grouped by task type: MCQ, OE, and Caption Generation.
}
\label{tab:task_field_descriptions}
\end{table*}

\subsection{Format conversion}
\label{format_concersion}

\begin{figure*}[h]
	\centering
	\includegraphics[width=1.0\textwidth]{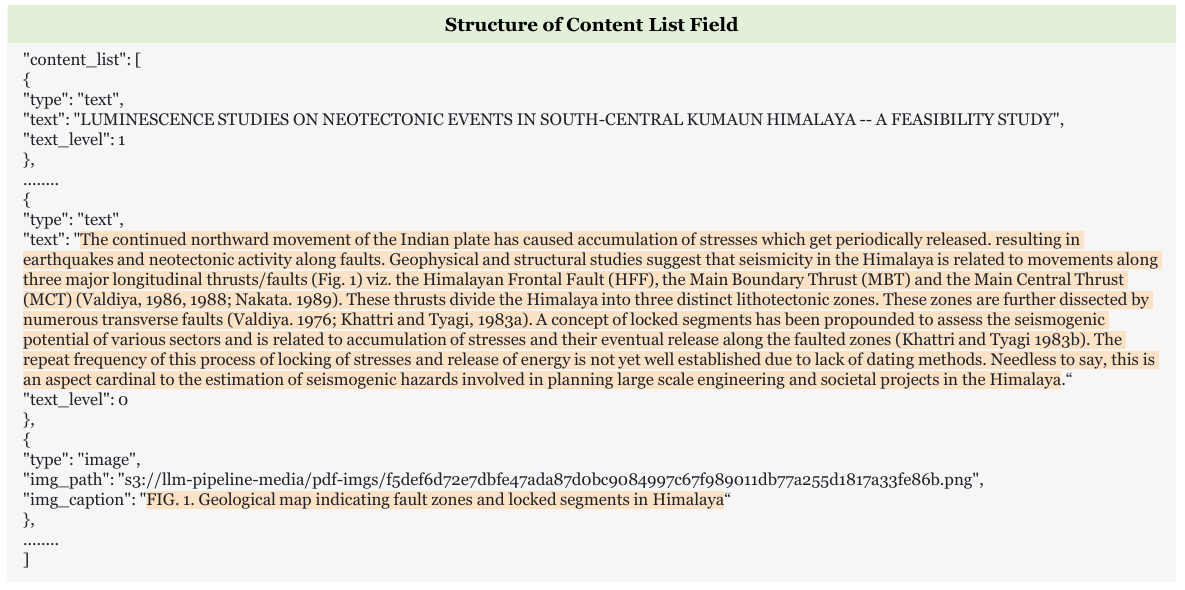}
        \caption{Examples of the content list field in a paper.}
        \label{mineru_case}
\end{figure*}

Specifically, our data source is a collection of papers gathered by OpenDataLab~\citep{he2024opendatalab} from online resources. These papers were processed using \textbf{MinerU}, which converted the textual content of the PDFs into JSON format and saved the images as PNG files. In Figure~\ref{mineru_case}, we present a portion of the content list from a processed PDF paper, highlighting the original caption (raw caption) of the image and the corresponding discussion section. It is evident that the discussion of figures in the paper contains substantial scientific reasoning, which is crucial for a comprehensive understanding of scientific figures.

\subsection{Paper Filtering}
\label{paper_filter}

To classify scientific papers into relevant Earth system categories, we employ a similarity-based approach using pre-trained sentence embeddings and cosine similarity. The process begins by generating embeddings for both the paper's title and predefined keywords using the pre-trained {all-MiniLM-L6-v2}~\citep{wang2020minilm} model, which captures the semantic meaning of textual data. First, we calculate the similarity between the paper's title and a set of general positive keywords, such as ``Earth,'' ``Earth system,'' ``hydrosphere,'' ``biosphere,'' ``lithosphere,'' ``atmosphere,'' and ``cryosphere.'' The cosine similarity is computed between the embedding of the paper's title and the embedding of the general positive keywords. If the similarity score is below a predefined threshold (0.2), the paper is excluded from further analysis, as it is deemed irrelevant to the sciences of the Earth system.

To further filter out irrelevant papers, we calculate the similarity between the paper's title and a set of general negative keywords, such as ``cell biology,'' ``virus,'' ``pharmaceuticals,'' ``chemistry,'' ``physics,'' and ``astronomy.'' If the similarity score exceeds a predefined threshold (0.1), the paper is excluded, as it is likely to belong to unrelated disciplines. For papers that pass the initial filtering, we calculate their similarity with predefined positive classification keywords for each Earth system category (e.g., hydrosphere, biosphere, lithosphere, atmosphere, cryosphere). Each category contains a list of domain-specific keywords. For example, the hydrosphere category includes keywords such as ``water cycle,'' ``ocean,'' ``rivers,'' ``lakes,'' and ``groundwater,'' while the biosphere category includes ``ecosystem,'' ``biodiversity,'' ``habitat,'' and ``species.'' The cosine similarity is computed between the paper's title and each keyword within a category, and the average similarity score for each category is calculated.

The category with the highest average similarity score is selected as the most relevant classification for the paper, provided the score exceeds a predefined threshold (0.15). To ensure robustness, we also calculate the similarity between the paper's title and negative classification keywords for each category. For example, the hydrosphere negative keywords include ``chemistry,'' ``universe,'' ``planets,'' and ``astronomy,'' while the biosphere negative keywords include ``cell biology,'' ``medicine,'' and ``pharmacology.'' The final relevance score for each category is computed as the difference between the positive and negative similarity scores, ensuring that papers with high relevance to unrelated fields are excluded. The final classification of a paper is determined by passing the general positive and negative keyword thresholds, identifying the category with the highest positive relevance score adjusted by subtracting the negative relevance score, and ensuring the adjusted relevance score exceeds the classification threshold (0.15). This approach allows us to systematically classify papers into Earth system categories while filtering out irrelevant content, leveraging semantic embeddings and cosine similarity to ensure that the classification is both accurate and interpretable.

\begin{table*}[h]
\centering
\resizebox{\textwidth}{!}{
\begin{tabular}{l c p{10cm}}
\toprule
{\bf Category} & {\bf Type} & {\bf Keywords} \\ 
\midrule

\texttt{General} & Positive & Earth, Earth system, hydrosphere, biosphere, lithosphere, atmosphere, cryosphere \\

\texttt{General} & Negative & cell biology, virus, pharmaceuticals, chemistry, physics, astronomy, food science, proteins, microbiology \\

\midrule

\texttt{Hydrosphere} & Positive & water cycle, ocean, rivers, lakes, groundwater, ice caps, aquifers, precipitation, evaporation, humidity \\

\texttt{Hydrosphere} & Negative & chemistry, universe, planets, astronomy, astrophysics, space, stars, galaxy, cosmology \\

\midrule

\texttt{Biosphere} & Positive & ecosystem, biodiversity, habitat, species, biomes, ecological balance, carbon cycle \\

\texttt{Biosphere} & Negative & cell biology, chemistry, medicine, pharmacology, microbiology, biochemistry, toxicology, pathology, clinical \\

\midrule

\texttt{Lithosphere} & Positive & earthquake, tectonic plates, earth's crust, minerals, rocks, soil, sediments, mountains, volcanoes, landforms, geological processes \\

\texttt{Lithosphere} & Negative & ancient texts, archaeology, culture, history, artifacts, civilization, prehistoric, mythology, anthropology \\

\midrule

\texttt{Atmosphere} & Positive & stratosphere, troposphere, weather, climate, greenhouse gases, ozone layer, air pressure, humidity, winds, carbon dioxide, temperature \\

\texttt{Atmosphere} & Negative & universe, galaxy, astronomy, astrophysics, space, stars, planets, cosmology, black holes, nebula, solar system \\

\midrule

\texttt{Cryosphere} & Positive & glaciers, ice sheets, sea ice, permafrost, snowpack, icebergs, frozen ground, climate change, albedo effect, polar regions \\

\texttt{Cryosphere} & Negative & frozen food, ice cream, refrigeration, freezing, cold storage, ice cubes, food preservation, chilling, frost \\

\bottomrule
\end{tabular}
}
\caption{
{Keywords for positive and negative classifications across different Earth system categories. The table includes general keywords as well as specific keywords for hydrosphere, biosphere, lithosphere, atmosphere, and cryosphere.}
}
\vspace{-3mm}
\label{tab:keywords_classification}
\end{table*}

\begin{figure*}[!t] 
    \centering
    \begin{tcolorbox}[
        colframe=black,
        colback=white,
        sharp corners,
        breakable,
        enhanced,
        boxrule=0.2mm,
        leftrule=0.2mm,
        rightrule=0.2mm,
        toprule=0.2mm,
        bottomrule=0.2mm,
        width=\textwidth
    ]
Analyze the provided image and classify it as an Earth observation image or not. 

Earth observation images include, but are not limited to:
\begin{itemize}[noitemsep, topsep=0pt]
    \item Remote sensing imagery,
    \item Atmospheric data visualizations,
    \item Aerial views of geographical features (e.g., rivers, urban landscapes),
    \item Weather-related images (e.g., precipitation maps, typhoon tracks),
    \item Cartographic representations.
\end{itemize}

Exclude images depicting:
\begin{itemize}[noitemsep, topsep=0pt]
    \item Biological entities (humans, plants, animals),
    \item Artificial objects (vehicles, device structures),
    \item Data visualizations (statistical charts, line graphs, scatter plots),
    \item Text-based content or blank images.
\end{itemize}

Output format:
\begin{itemize}[noitemsep, topsep=0pt]
    \item Return "1" if the image is an Earth observation image.
    \item Return "0" if the image does not meet the Earth observation criteria.
\end{itemize}

Provide only the numerical output (1 or 0) without any additional text or explanation.
\end{tcolorbox}
    \caption{Prompt for retaining Earth observation images.} 
    \label{box:earth_obs} 
\end{figure*}

\subsection{Image Filtering}
\label{image_filter}

Next, we further filtered the images in these 103,108 papers. Our goal was to retain Earth observation images, as our task focuses on evaluating the model's ability to understand and reason about scientific phenomena in Earth sciences. These images include various types of visual data, such as those representing geophysical processes, atmospheric phenomena, geographic features, weather patterns, and cartographic representations.

To achieve this, we employed a systematic filtering pipeline based on the \texttt{Qwen2-VL-7B-Instruct} model. The filtering process was guided by a carefully designed prompt, which instructed the model to classify each image as either an Earth observation image or not. Specifically, the prompt defined Earth observation images as those depicting remote sensing imagery, atmospheric data visualizations, aerial views of geographical features (e.g., rivers, urban landscapes), weather-related images (e.g., precipitation maps, typhoon tracks), and cartographic representations. Conversely, the prompt explicitly excluded images containing biological entities (e.g., humans, plants, animals), artificial objects (e.g., vehicles, device structures), data visualizations (e.g., statistical charts, line graphs, scatter plots), text-based content, or blank images.

The filtering process was implemented as follows: for each image, the model was provided with both the image and the prompt, and it generated a binary output (``1'' for Earth observation images and ``0'' otherwise). To ensure robustness, the model's output was validated through multiple sampling attempts with slight variations in generation parameters (e.g., temperature). If the model consistently classified an image as ``1,'' it was retained; otherwise, it was discarded. This iterative and robust classification approach allowed us to minimize false positives and negatives in the filtering process. After this step, we retained around 83K papers, which contained images classified as Earth observation images. These filtered images form the basis for subsequent analysis and evaluation of the model's capabilities in understanding and reasoning about Earth science phenomena. Prompt for retaining Earth observation images are shown in Figure~\ref{box:earth_obs}.

\subsection{Content Filtering}
\label{content_filter}
To construct our benchmark, which requires generating VQA tasks supported by the content of the papers, we ensured that the selected figures not only had captions but were also discussed in detail within the text of the papers. Using regular expressions, we extracted the figure numbers and identified corresponding discussions in the main body of the papers. Figures with discussions exceeding two sentences were included in the final dataset. Finally, we selected 64,560 papers, resulting in a total of 289,891 figures for further processing.

\begin{figure}[h]
    \centering
    \includegraphics[width=0.45\textwidth]{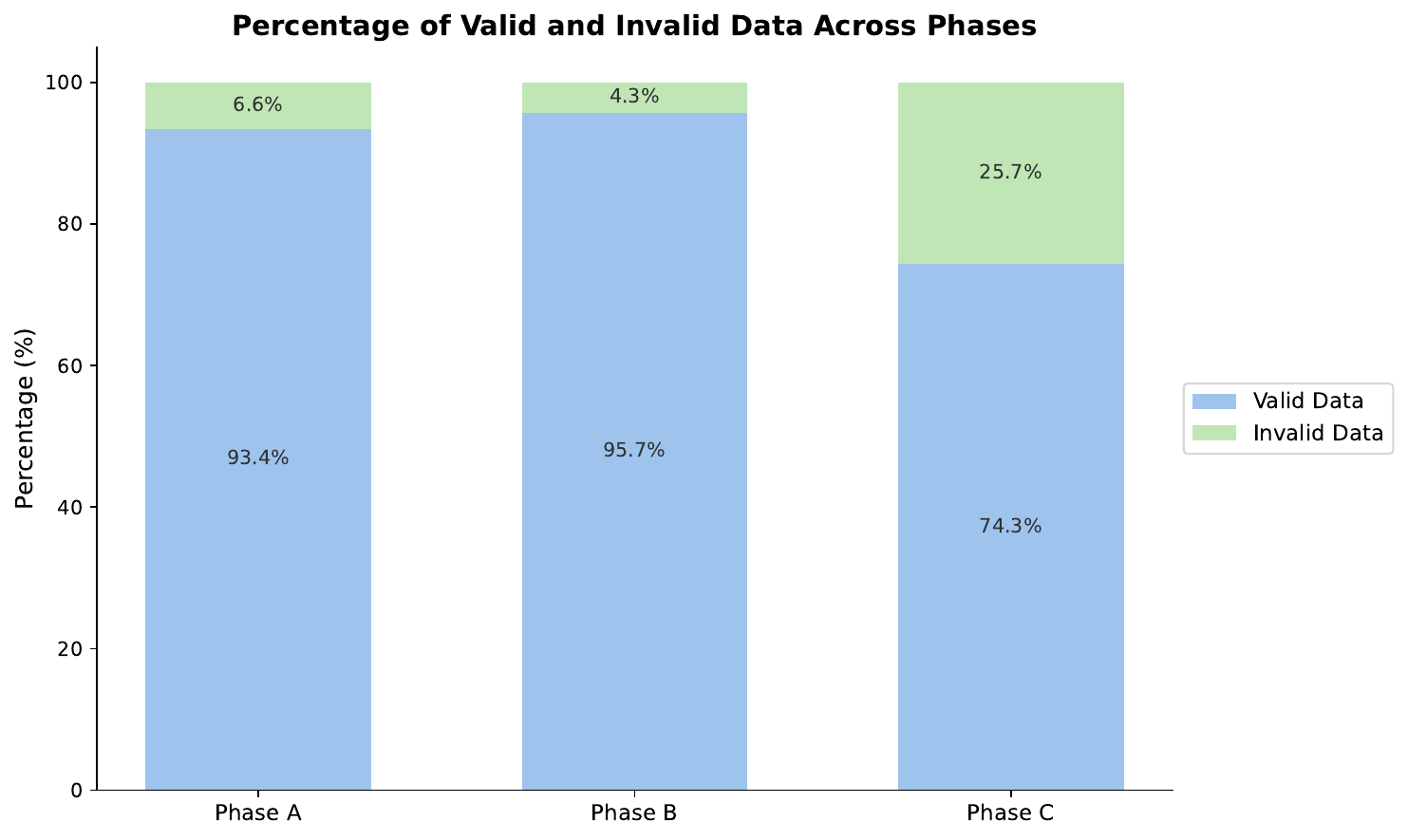}
    \caption{Proportion of valid and invalid data after manual screening across different phases. Phase B, or Specialized VQA, exhibits the highest quality.}
    \label{phase_percentage}
\end{figure}

\subsection{Prompt Designer for \modelname}
\label{prompt_design}

The prompt was used to generate a refined caption are shown in Figure~\ref{box:refined_cap}.

\begin{figure*}[h] 
    \centering
    \begin{tcolorbox}[
        colframe=black,
        colback=white,
        sharp corners,
        enhanced,
        boxrule=0.2mm,
        width=\textwidth
    ]
    You are an expert assistant in scientific image analysis and caption generation. Your task is to rewrite or generate a new, detailed caption for the provided figure using the original caption and only the sentences or information from the Relevant Content that are directly associated with this figure.
    
    \medskip 
    
    \textbf{Please strictly follow these guidelines:}
    \begin{itemize}[noitemsep, topsep=2pt, parsep=0pt] 
        \item Assume the figure does not reference or depend on other figures in the document.
        \item Exclude any mention of other figures, their content, or references in the caption.
        \item If subfigures are present, provide specific descriptions for each subfigure accordingly. Otherwise, assume it represents a single figure.
        \item The new caption must be detailed, precise, and include only the relevant details from the provided content.
    \end{itemize}

    \medskip 

    \textbf{Inputs for caption generation:}
    \begin{itemize}[noitemsep, topsep=2pt, parsep=0pt]
        \item Original Caption: \{caption\}
        \item Relevant Content: \{content\}
    \end{itemize}

    \medskip 

    Now write a detailed, high-quality caption for this figure below:
    \end{tcolorbox}
    
    \caption{Prompt for generating refined captions.} 
    \label{box:refined_cap} 
\end{figure*}

The prompt was used to generate diverse VQAs are shown in Figure~\ref{box:vqas}.

\begin{figure*}[!t] 
    \centering
    \begin{tcolorbox}[
        colframe=black,
        colback=white,
        sharp corners,
        enhanced,
        boxrule=0.2mm,
        width=\textwidth
    ]
You are an advanced AI model specialized in generating high-quality Visual Question Answering (VQA) tasks. Your role is to generate a diverse set of VQA questions, answers, and reasoning chains based on the provided visual input (a figure) and its captions.

\section*{Definitions:}
\begin{enumerate}[noitemsep, topsep=0pt]
    \item \textbf{Figure:} A scientific or illustrative figure provided as the primary visual input. Test-takers will analyze this image to answer the questions.
    \item \textbf{Caption:} A concise summary describing key aspects of the Figure.
    \item \textbf{Supplementary:} In-depth information (e.g., summarized expert insight, detailed analysis, or background knowledge) that you can use to assist in designing advanced and meaningful questions. However, test-takers cannot access this information.
\end{enumerate}

\section*{Input Information Provided:}
\begin{itemize}[noitemsep, topsep=0pt]
    \item \textbf{Caption:} \{raw caption\}
    \item \textbf{Supplementary:} \{refined caption\}
\end{itemize}

\section*{Task Instructions:}

\subsection*{1. Use of Input Sources:}
\begin{itemize}[noitemsep, topsep=0pt]
    \item Ensure that no question can be answered entirely using Caption without observations.
    \item \textbf{Supplementary Usage:} The correct answers are encouraged to be derived from...
\end{itemize}

\subsection*{2. Question Types:}
\begin{itemize}[noitemsep, topsep=0pt]
    \item \textbf{Multiple Choice Questions (MCQs):} At least \textbf{2} questions must be of this type, with 4 distinct options (A-D) and one correct answer.
    \item \textbf{Open-Ended Questions:} At least \textbf{2} questions must be open-ended, requiring concise and precise answers (no more than 4 words).
\end{itemize}

\subsection*{3. Reasoning Chains:}
\begin{itemize}[noitemsep, topsep=0pt]
    \item For every question, you must include a reasoning chain. The chain explains the logical process by which the correct answer can be determined.
    \item The reasoning chain must:
\end{itemize}

\subsection*{4. Output Structure:}
The output must be written in \textbf{JSON format}

\subsection*{5. Task Guidelines:}
\begin{enumerate}[noitemsep, topsep=0pt]
    \item Questions that are grounded in the Supplementary context are highly encouraged. These questions should require the test-taker to refer to in-depth knowledge and insights not immediately visible in the Figure or Caption.
    \item Avoid referencing the Supplementary in any question and \texttt{reasoning\_chain} (e.g., "According to the Supplementary" or "The Supplementary states").
\end{enumerate}

Provide your response below:
    \end{tcolorbox}
    
    \caption{Prompt for generating VQAs.} 
    \label{box:vqas} 
\end{figure*}

\section{Multi-Agent Voting}
\label{agnet-voting}

\subsection{Prompt}

The prompt was used to generate a normal answer for MCQ are shown in Figure~\ref{box:normal_ans}.

\begin{figure*}[!t] 
    \centering
    \begin{tcolorbox}[
        colframe=black,
        colback=white,
        sharp corners,
        enhanced,
        boxrule=0.2mm,
        width=\textwidth
    ]
You are tasked with answering a multiple-choice question about the given input image.

\section*{Instructions:}
\begin{enumerate}[noitemsep, topsep=0pt]
    \item Carefully analyze the input image and the provided query.
    \item Based on the image, select the correct option (e.g., 'A', 'B', 'C') or directly state the correct option content.
    \item Provide reasoning explaining how to derive the correct answer.
\end{enumerate}

\section*{Input:}
\begin{itemize}[noitemsep, topsep=0pt]
    \item \textbf{Query:} \{query\}
\end{itemize}

\section*{Output Format:}
The output must be written in \textbf{JSON format} using the structure below:
\begin{verbatim}
{
    "answer": "Correct option or short answer",
    "Explanation": "Explaining how to derive correct answer."
}
\end{verbatim}

    \end{tcolorbox}
    
    \caption{Prompt for generating normal answers for MCQs.} 
    \label{box:normal_ans} 
\end{figure*}

The prompt was used to generate a enhanced answer for MCQ are shown in Figure~\ref{box:enhanced_ans}.

\begin{figure*}[h] 
    \centering
    \begin{tcolorbox}[
        colframe=black,
        colback=white,
        sharp corners,
        enhanced,
        boxrule=0.2mm,
        width=\textwidth
    ]
You are tasked with answering a multiple-choice question about the given input image.

\section*{Input:}
\begin{itemize}[noitemsep, topsep=0pt]
    \item \textbf{Question:} \{question\}
    \item \textbf{Refined Caption:} \{caption\}
\end{itemize}

\section*{Instructions:}
\begin{enumerate}[noitemsep, topsep=0pt]
    \item Carefully analyze the input image and its caption.
    \item Based on the image and caption, select the correct option (e.g., 'A', 'B', 'C') or directly state the correct option content.
\end{enumerate}

\section*{Output Format:}
The output must be written in \textbf{JSON format} using the structure below:
\begin{verbatim}
{
    "answer": "Correct option or short answer",
    "Explanation": "Explaining how to derive correct answer."
}
\end{verbatim}
    \end{tcolorbox}
    
    \caption{Prompt for generating enhanced answers for MCQs.} 
    \label{box:enhanced_ans} 
\end{figure*}

\begin{figure*}[h]
	\centering
	\includegraphics[width=1.0\textwidth]{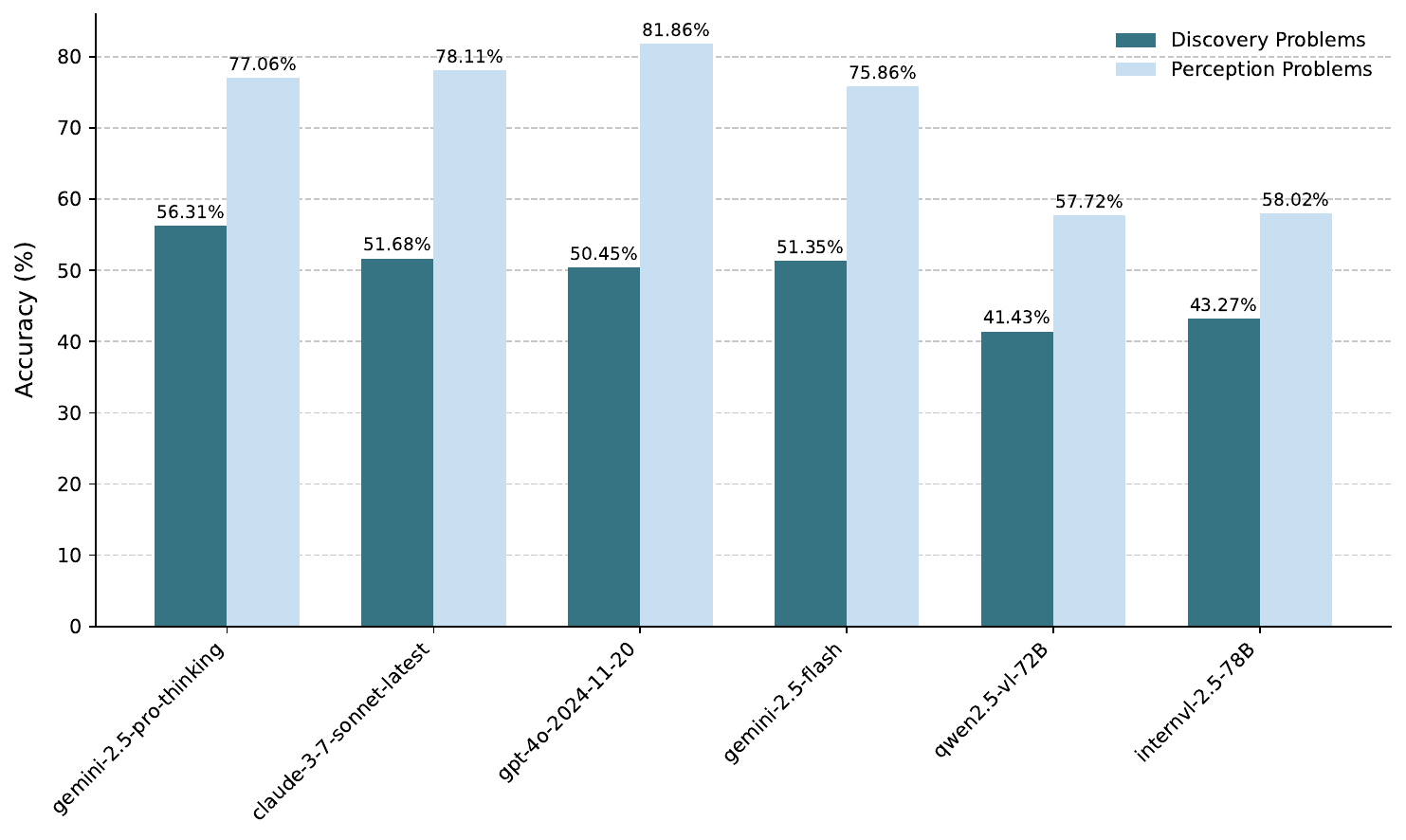}
        \caption{Models' accuracy on \qatype{} and perception problems.}
        \label{vqa_accuracy}
\end{figure*}

\subsection{Example of Different Levels of Questions}
\label{level_example}

\begin{figure*}[h]
	\centering
	\includegraphics[width=1.0\textwidth]{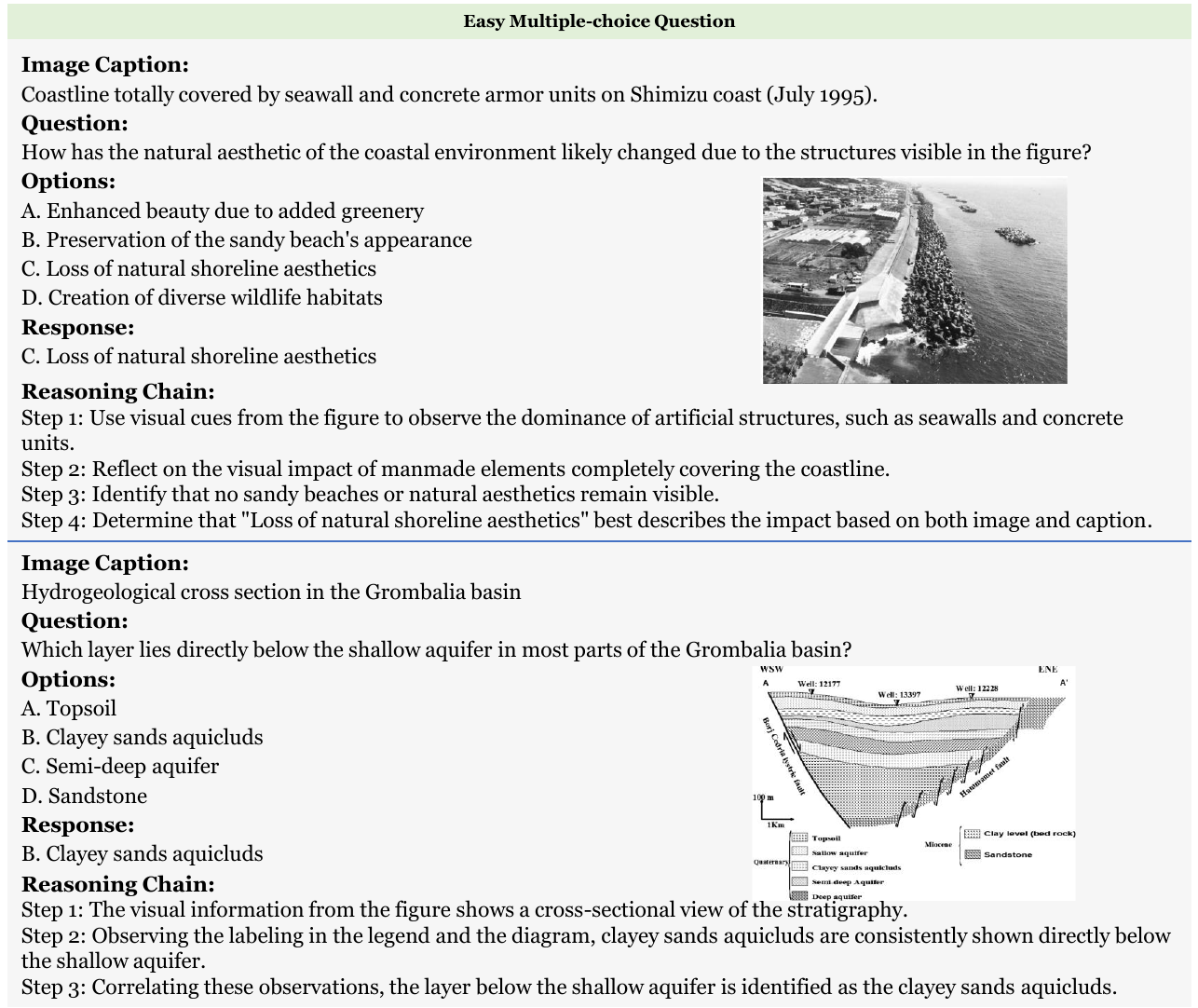}
        \caption{An example of easy multiple-choice VQA.}
        \label{easy_case}
\end{figure*}

\begin{figure*}[h]
	\centering
	\includegraphics[width=0.9\textwidth]{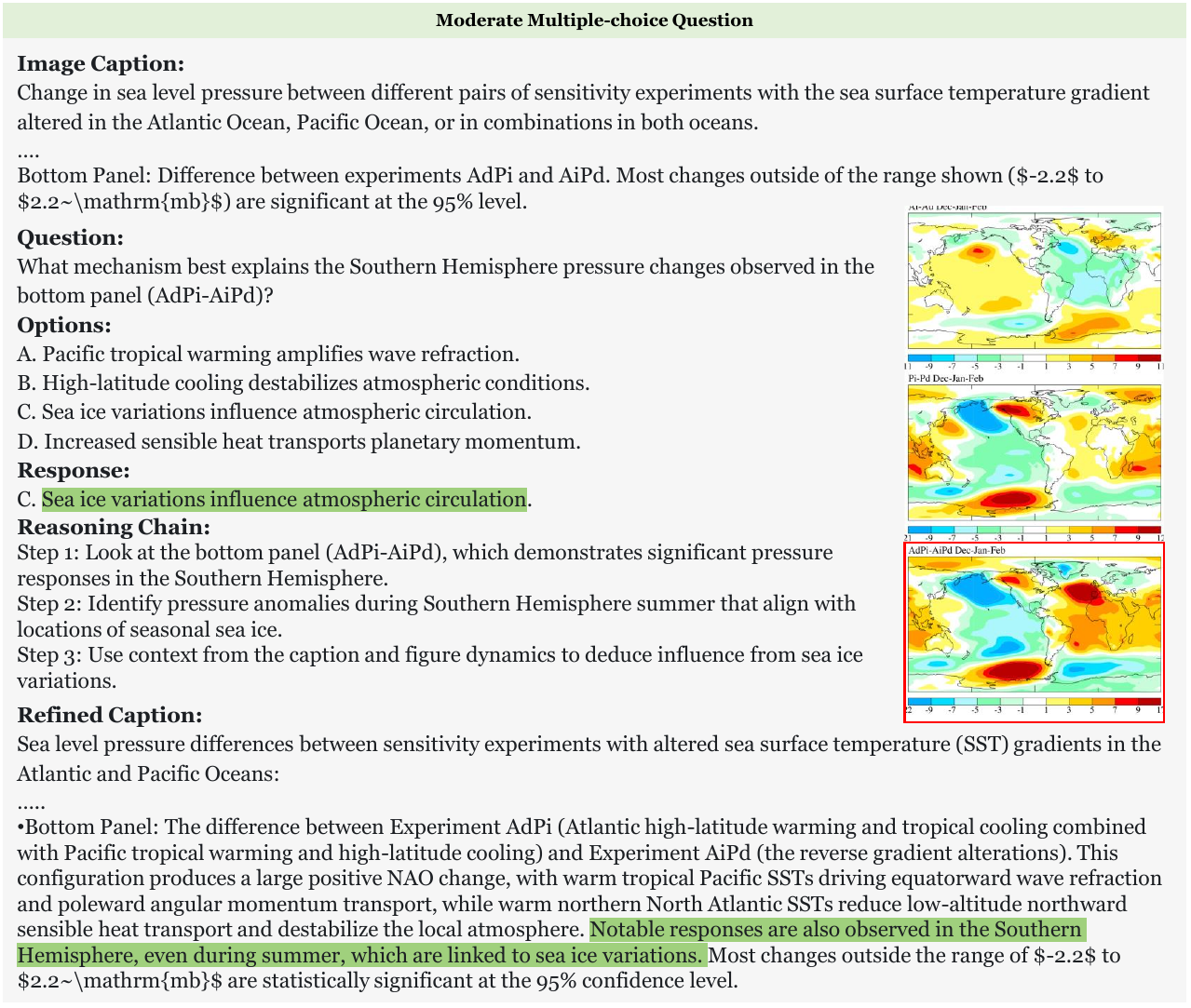}
        \caption{An example of specialized multiple-choice VQA.}
        \label{moderate_case}
\end{figure*}

\begin{figure*}[h]
	\centering
	\includegraphics[width=0.9\textwidth]{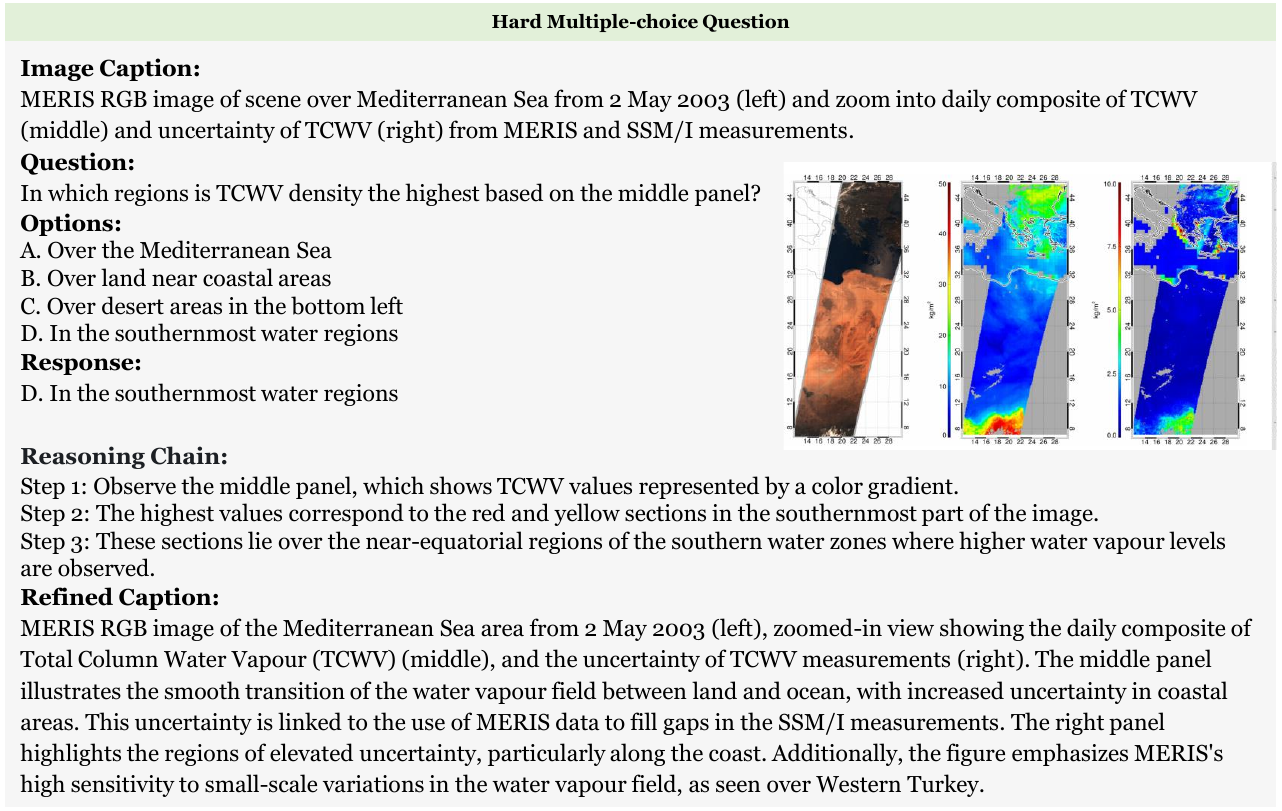}
        \caption{An example of hard multiple-choice VQA.}
        \label{hard_case}
\end{figure*}

Figure~\ref{easy_case} illustrates an example of a simple problem in multi-agent voting, while Figure~\ref{moderate_case} presents an example of a domain-specific problem, and Figure~\ref{hard_case} demonstrates an example of a challenging problem. The most notable distinction lies in the varying levels of perceptual ability required by the model: simple and challenging problems primarily differ in the model's ability to perceive and interpret images, whereas domain-specific problems emphasize the model's knowledge in specialized fields. Additionally, the answers to domain-specific questions are often supported by evidence found in the "refined caption" field provided in the paper.

To construct the benchmark datasets, we sampled data from the multi-agent automated filtering process as follows: 900 questions from Phase A, 1800 questions from Phase B, and 300 questions from Phase C were selected to form the multiple-choice question (MCQ) set, while 500 questions from Phase A and 1000 questions from Phase B were selected to form the open-ended question set. All sampled data were subsequently validated by experts to ensure accuracy and quality.

\section{Expert Validation}
\label{expert_val}

\subsection{Details}
We recruited annotators with a background in Earth sciences and a master's degree through an annotation company to label the data. The annotated dataset consists of 3,000 MCQs and 1,500 open-ended QAs. We provided the annotators with figures, queries, reasoning chains, and our processed refined captions to assist them in evaluating whether the provided answers were reasonable. For questions where the answers could not be found in the refined captions, the annotators were required to use their own knowledge to determine the correctness of the answers. If they were unable to make a judgment, such questions were discarded to ensure that the filtered dataset contained only accurate and complete questions. The tasks assigned to the annotators are described below:

The evaluation framework categorizes questions based on several criteria. First, the \textbf{Image Type of Reasoning Required} distinguishes between questions involving a single image, where the input consists of just one image, and those with a \textbf{Single-image focus}, where multiple images are present but the question pertains to one specific image. Additionally, \textbf{Multi-image reasoning} questions require comparing or reasoning across multiple images.

Next, the \textbf{Type of Scientific Question} is considered. \textbf{Perception Questions} are those where answers can be derived through basic observation, such as identifying position or color. These questions do not have answers in the refined captions and require manual evaluation of their validity. In contrast, \textbf{Reasoning Questions} necessitate domain-specific knowledge for answering, and annotators must verify if the answer can be derived from the refined caption field.

The \textbf{Completeness of Questions} is another criterion, where questions are classified as \textbf{Complete} if all necessary information is provided in the question or image, and \textbf{Incomplete} if missing information makes it difficult or impossible to answer.

Finally, the \textbf{Correctness of Questions} assesses whether the provided answer is accurate, categorizing them as \textbf{Correct} or \textbf{Incorrect} based on the accuracy of the answer. After manual screening, 216 invalid entries were identified in the MCQ task, and 89 invalid entries were found in the open-ended task. To evaluate the effectiveness of our multi-agent filtering process, we conducted a statistical analysis of the three phases of data. In Phase A, 59 out of 900 sampled questions were deemed invalid after manual review; in Phase B, 80 out of 1800 questions were invalid; and in Phase C, 77 out of 300 questions were invalid. These results demonstrate the utility of the initial model-based filtering: questions supported by refined captions and correctly answered by most models (Phases A and B) tend to be of higher quality, while questions filtered in Phase C exhibit lower quality.

\section{Classification of Research Problems in Earth Sciences}
\label{classification}

Under the framework of the five major spheres, we further categorized the generated research problems into specific academic disciplines according to a standardized classification system. Within the broad category of Earth Sciences, we refined the classification into detailed sub-disciplines or sub-fields. The classification process involves three main steps: first, identifying the primary sphere to which the research problem belongs, selecting from eight major disciplines (referred to as primary spheres), including Atmospheric Sciences, Ecology and Biosciences, Hydrology, Oceanography, Geology, Geography, Solid Earth Geophysics, and Polar Science. Second, the classification is further refined by selecting the most appropriate sub-discipline or sub-field from a detailed hierarchy. Third, for interdisciplinary problems, the primary classification is clearly stated, and any relevant secondary classifications are noted. This hierarchical approach ensures a systematic and precise categorization of research problems, enabling a deeper understanding of their academic and scientific context.

\textbf{Summary of Classification:} The classification system includes a total of 8 first-level disciplines and 66 second-level disciplines. Each research problem is assigned to one of the primary disciplines and further refined into a specific sub-discipline based on its characteristics and context.

\subsection*{Classification Hierarchy}

\begin{description}
    \item[Atmospheric Sciences:] 
    Atmospheric Chemistry, Meteorology, Climatology, Hydrometeorology, Paleoclimatology, Atmospheric Physics, Numerical Weather Prediction and Simulation, Atmospheric Remote Sensing.

    \item[Ecology and Biosciences:] 
    Regional Ecology, Population Ecology, Community Ecology, Ecosystem Ecology, Ecological Engineering, Restoration Ecology, Landscape Ecology, Aquatic Ecology and Limnological Ecology, Biogeochemistry, Biogeography.

    \item[Hydrology:] 
    Hydrology, Hydrogeology, Limnology, River Hydrology and Estuarine Hydrology, Groundwater Hydrology, Regional Hydrology, Ecohydrology, Hydrological Physics, Hydrological Geography, Hydrological Meteorology, Hydrological Measurement, Hydrological Cartography.

    \item[Oceanography:] 
    Ocean Chemistry, Ocean Physics, Ocean Biology, Ocean Geology, Remote Sensing Oceanography, Environmental Oceanography, Marine Resources Science.

    \item[Geology:] 
    Economic Geology, Engineering Geology, Environmental Geology, Quaternary Geology, Sedimentology, Stratigraphy, Paleogeography, Volcanology, Mineralogy and Petrology, Regional Geology, Remote Sensing Geology.

    \item[Geography:] 
    Physical Geography, Human Geography, Regional Geography, Urban Geography, Tourism Geography, World Geography, Historical Geography, Geomorphology, Biogeography, Chemical Geography, Other Disciplines in Geography.

    \item[Solid Earth Geophysics:] 
    Geodynamics, Seismology, Geomagnetism, Gravimetry, Geoelectricity, Geothermal Science, Tectonophysics, Exploration Geophysics, Computational Geophysics, Experimental Geophysics, Other Disciplines in Solid Earth Geophysics.

    \item[Polar Science:] 
    Polar Ecology, Polar Oceanography, Glaciology, Permafrost Science, Polar Climate Science.
\end{description}

\begin{table*}[ht]
\centering
\caption{
\textbf{Top Sub-disciplines in Various Scientific Subjects.}
The table lists the top three sub-disciplines by count within each major scientific subject.
}
\resizebox{\textwidth}{!}{%
\begin{tabular}{l|lcc|lcc|lcc}
\toprule
{\textbf{Subject}} & \multicolumn{2}{c}{\textbf{Top 1 Sub-subject}} & &\multicolumn{2}{c}{\textbf{Top 2 Sub-subject}} & &\multicolumn{2}{c}{\textbf{Top 3 Sub-subject}} \\
\midrule

Hydrology & River Hydrology and Estuarine Hydrology & 805 & & Groundwater Hydrology & 790 & & Limnology & 439 & \\
Ecology and Biosciences & Aquatic Ecology and Limnological Ecology & 562 & & Landscape Ecology & 298 & & Ecosystem Ecology & 280 & \\
Geology & Sedimentology & 1068 & & Quaternary Geology & 298 & & Structural Geology & 215 & \\
Solid Earth Geophysics & Seismology & 845 & & Tectonophysics & 343 & & Exploration Geophysics & 74 & \\
Geography & Physical Geography & 1575 & & Urban Geography & 76 & & Geomorphology & 40 & \\
Polar Science & Glaciology & 352 & & Polar Climate Science & 46 & & Permafrost Science & 31 & \\
Atmospheric Sciences & Meteorology & 920 & & Climatology & 619 & & Atmospheric Remote Sensing & 159 & \\
Oceanography & Ocean Physics & 698 & & Ocean Geology & 163 & & Environmental Oceanography & 104 & \\

\bottomrule
\end{tabular}%
}
\label{tab:sub_disciplines}
\end{table*}

\section{MLLMs' versions}
\label{baselines}

\begin{table*}[ht]
\scriptsize
\centering
\caption{Evaluated MLLMs in our experiments with their versions or Huggingface model paths.}
\resizebox{\textwidth}{!}{
\begin{tabular}{l|l}
\toprule
\rowcolor{Gray} \multicolumn{2}{c}{\textit{\textbf{Open-source Models}}} \\
\textbf{Model} & \textbf{Model path} \\
\midrule
Qwen2.5-VL-7B & \url{https://huggingface.co/Qwen/Qwen2.5-VL-7B-Instruct} \\
Qwen2.5-VL-32B & \url{https://huggingface.co/Qwen/Qwen2.5-VL-32B-Instruct} \\
Qwen2.5-VL-72B & \url{https://huggingface.co/Qwen/Qwen2.5-VL-72B-Instruct} \\
InternVL2.5-8B & \url{https://huggingface.co/OpenGVLab/InternVL2_5-8B} \\
InternVL2.5-78B & \url{https://huggingface.co/OpenGVLab/InternVL2_5-78B} \\
InternVL3-78B & \url{https://huggingface.co/OpenGVLab/InternVL3-78B} \\
LLaVA-onvision-72B & \url{https://huggingface.co/llava-hf/llava-onevision-qwen2-72b-ov-hf} \\
Llama3.2-90B-Vision & \url{https://huggingface.co/meta-llama/Llama-3.2-90B-Vision} \\
DeepSeek-VL2 & \url{https://huggingface.co/deepseek-ai/deepseek-vl2} \\
Intern-S1-mini & \url{https://huggingface.co/internlm/Intern-S1-mini} \\
\midrule
\rowcolor{Gray} \multicolumn{2}{c}{\textit{\textbf{Proprietary Models}}} \\
\textbf{Model} & \textbf{Model versioning} \\
\midrule
GPT-4o & \texttt{GPT-4o-20} \\
Gemini-2.5-Pro-Thinking & \texttt{gemini-2.5-pro-preview-05-06} \\
Gemini-2.5-Flash & \texttt{Gemini-2.5-Flash-preview-04-17} \\
Gemini-2.5-Flash-Thinking & \texttt{Gemini-2.5-Flash-preview-04-17} \\
Claude-3.7-Sonnet & \texttt{Claude-3.7-Sonnet-20250219} \\
Claude-3.5-Haiku & \texttt{claude-3-5-haiku-20241022} \\
GPT-4o-mini & \texttt{GPT-4o-mini-2024-07-18} \\
\bottomrule
\end{tabular}
}
\label{tab:model_path}
\end{table*}

For open-source models, we use vllm~\citep{kwon2023efficient} for local testing; for proprietary models, we conduct tests via API calls. The download paths for specific models and the versions of models accessed via API are provided in Figure~\ref{tab:model_path}.

\section{Evaluation Metrics}
\label{metrics}

\subsection{MLLM-based Metrics}
\label{mllm_metrics}
Following G-Eval~\citep{liu2023g}, we utilize MLLM (Qwen2.5-VL-72B) with a specialized prompt to compute a factual scientific score.
For the captioning task, we define a Cap-Eval score ranging from 1 to 5, where higher scores indicate better caption quality. For the open-ended QA task, we introduce OE-Eval, which evaluates the reasonableness of generated answers using a binary 0/1 scoring system.

The prompt was used for Cap-Eval are shown in Figure~\ref{box:eva_cap}.

\begin{figure*}[h] 
    \centering
    \begin{tcolorbox}[
        colframe=black,
        colback=white,
        sharp corners,
        enhanced,
        boxrule=0.2mm,
        width=\textwidth
    ]
Evaluate the quality of a generated caption for a geoscience research paper figure or image.

\section*{Evaluation Criteria:}
\begin{enumerate}[noitemsep, topsep=0pt]
    \item \textbf{Scientific Accuracy:} Does the generated caption accurately describe the scientific content of the figure or image?
    \item \textbf{Clarity and Coherence:} Is the caption well-structured, logically organized, and easy to understand?
    \item \textbf{Relevance and Completeness:} Does the caption provide all necessary information to understand the figure or image?
\end{enumerate}

\section*{Evaluation Steps:}
\begin{enumerate}[noitemsep, topsep=0pt]
    \item Compare the \textbf{Generated Caption} to the \textbf{Standard Caption}. Assess whether the generated caption aligns with the scientific content and intent of the standard caption.
    \item Assign a score for coherence on a scale of 1 to 5, where 1 is the lowest and 5 is the highest, based on the Evaluation Criteria.
\end{enumerate}

\section*{Input:}
\begin{itemize}[noitemsep, topsep=0pt]
    \item \textbf{Standard Caption:} \{response\}
    \item \textbf{Generated Caption:} \{generated\_caption\}
\end{itemize}

\section*{Important Instructions:}
\begin{itemize}[noitemsep, topsep=0pt]
    \item Only output the score in the specified JSON format.
    \item Do not provide any explanations, comments, or additional text.
\end{itemize}

\section*{Output Format:}
The output must be written in \textbf{JSON format} using the structure below:
\begin{verbatim}
{
    "score": 1-5
}
\end{verbatim}
    \end{tcolorbox}
    
    \caption{Prompt for evaluating the quality of generated captions.} 
    \label{box:eva_cap} 
\end{figure*}

The prompt was used for OE-Eval are shown in Figure~\ref{box:eva_oe}.

\begin{figure*}[h] 
    \centering
    \begin{tcolorbox}[
        colframe=black,
        colback=white,
        sharp corners,
        enhanced,
        boxrule=0.2mm,
        width=\textwidth
    ]
You are tasked with evaluating the correctness of a generated answer to an open-ended question about a given input image.

\section*{Input:}
\begin{itemize}[noitemsep, topsep=0pt]
    \item \textbf{Question:} \{query\}
    \item \textbf{Refined Caption:} \{refined caption\}
    \item \textbf{Standard Answer:} \{response\}
    \item \textbf{Generated Answer:} \{generated\_answer\}
\end{itemize}

\section*{Instructions:}
\begin{enumerate}[noitemsep, topsep=0pt]
    \item Based on the refined caption, question, and standard answer, determine if the generated answer is correct.
    \item Only output the determination in the specified JSON format.
    \item Do not provide any explanations, comments, or additional text.
\end{enumerate}

\section*{Output Format:}
The output must be written in \textbf{JSON format} using the structure below:
\begin{verbatim}
{
    "is_correct": true or false
}
\end{verbatim}
    \end{tcolorbox}
    
    \caption{Prompt for evaluating the quality of generated answers to open-ended questions.} 
    \label{box:eva_oe} 
\end{figure*}

To further establish the correlation between LLMs and human judgment specifically in the domain of Earth Science VQA, we conducted a human evaluation with four Ph.D. candidates specializing in Earth sciences. They scored a random sample of 160 questions from our MSEarth Open Ended benchmark. The models evaluated included Gemini-2.5-Flash, GPT-4o, InternVL3-78B and QwenVL2.5-72B. Our inter-annotator agreement, measured by Krippendorff’s alpha, is 69.5. Following LAVE~\citep{manas2024improving}, in order to assess the validity of OE-Eval, we calculated its correlation with human judgment using Spearman’s rank correlation coefficients. We derive a single “quality” score from the 4 binary ratings (correct/incorrect) per answer as follows: 1.0 if at least 3 annotators rate the answer as correct, 0.5 if only 2 did so, and 0.0 otherwise.


From the table's results, OE-Eval demonstrates a higher consistency with human judgment compared to all the considered baselines.

\subsection{Similarity-based Metrics}

In cases where some models fail to strictly follow instructions and only output the correct answer, resulting in regular expression matching failures, we use the all-MiniLM-L6-v2 model~\citep{wang2020minilm} to calculate the similarity between the model's output and each option. The option with the highest similarity is then selected as the model's answer.

\section{Detailed MSEarth-MCQ Results}
\label{detailed_MCQ}

The inputs to our model, \modelname{}, consist of images, questions, and the original captions. The original captions provide contextual information about the images, such as the meanings of specific symbols. Therefore, we conducted tests on different models to evaluate their performance with and without the original captions.

\begin{table*}[ht]
\scriptsize
\centering
\caption{{Accuracies (\%) of different models on multiple-choice questions.} The best results are highlighted in bold, with the second-best underlined. OC: original caption.}
\resizebox{\textwidth}{!}{
\begin{tabular}{lccccccc}
\toprule

\multirow{2}{*}{\textbf{Model}} & {\textbf{Input}} & \multicolumn{3}{c}{\textbf{Image-Type}} & \multicolumn{2}{c}{\textbf{Task Type}} & \multicolumn{1}{c}{\textbf{Overall}} \\

 & OC & \textbf{\textsc{Single}} & \textbf{\textsc{Multi}} & \textbf{\textsc{Cross}}  & \textbf{\textsc{\qatype{}}} & \textbf{\textsc{Percept}} & \textbf{\textsc{ACC}} \\
\midrule

\rowcolor{Gray} \multicolumn{8}{c}{\textit{\textbf{Open-source Models}}} \\

LLaVA-onvision-72B & \xmark & 49.40 & 45.52 & 41.10 & 41.92 & 61.86 & 46.69 \\
Qwen2.5-VL-7B & \xmark & 39.12 & 35.65 & 39.18 & 37.27 & 38.98 & 37.68 \\
Qwen2.5-VL-32B & \xmark & 42.07 & 39.78 & 40.00 & 37.03 & 52.92 & 40.84 \\
Qwen2.5-VL-72B & \xmark & 47.65 & 43.30 & 43.84 & 41.43 & 57.72 & 45.33 \\
InternVL2-8B & \xmark & 35.94 & 34.11 & 32.33 & 34.25 & 36.13 & 34.70 \\
InternVL2.5-78B & \xmark & 48.13 & 45.88 & 45.21 & 43.27 & 58.02 & 46.80 \\
InternVL3-78B & \xmark & 51.95 & 44.85 & 45.75 & 44.54 & 59.67 & 48.17 \\
Llama3.2-90B-Vision & \xmark & 44.30 & 40.64 & 36.16 & 38.64 & 51.42 & 41.70 \\
DeepSeek-VL2 & \xmark & 45.42 & 42.70 & 46.85 & 43.74 & 46.78 & 44.47 \\

LLaVA-onvision-72B & \cmark & 53.55 & 49.48 & 47.95 & 46.58 & 65.52 & 51.11 \\
Qwen2.5-VL-7B & \cmark & 47.65 & 44.07 & 37.53 & 40.53 & 58.47 & 44.83 \\
Qwen2.5-VL-32B & \cmark & 52.59 & 46.99 & 43.84 & 42.47 & 70.16 & 49.10 \\
Qwen2.5-VL-72B & \cmark & 52.11 & 50.43 & 46.30 & 44.40 & 70.46 & 50.65 \\
InternVL2-8B & \cmark & 44.86 & 43.99 & 38.36 & 38.97 & 58.47 & 43.64 \\
InternVL2.5-78B & \cmark & 53.23 & 49.74 & 44.38 & 43.17 & 74.21 & 50.61 \\
InternVL3-78B & \cmark & 57.53 & 51.37 & 45.48 & 47.00 & 73.61 & 53.38 \\
Llama3.2-90B-Vision & \cmark & 45.98 & 40.46 & 38.90 & 38.26 & 56.97 & 42.74 \\
DeepSeek-VL2 & \cmark & 52.43 & 49.23 & 44.66 & 46.06 & 62.82 & 50.07 \\
\midrule
\rowcolor{Gray}
\multicolumn{8}{c}{\textit{\textbf{Proprietary Models}}} \\

Gemini-2.5-Flash & \cmark & 58.33 & 54.55 & \underline{53.42} & 49.98 & 75.56 & 56.11 \\
Gemini-2.5-Flash-Thinking & \cmark & {60.64} & 54.64 & 53.70 & \underline{51.35} & 75.86 & 57.22 \\
Gemini-2.5-Pro-Thinking & \cmark & \textbf{64.78} & \textbf{59.36} & {55.34} & \textbf{56.31} & 77.06 & \textbf{61.28} \\
Claude-3.5-Haiku & \cmark & 49.48 & 47.16 & 42.47 & 42.18 & 64.77 & 47.59 \\
Claude-3.7-Sonnet & \cmark & 59.52 & \underline{56.53} & \textbf{57.53} & {51.68} & \underline{78.11} & \underline{58.01} \\
GPT-4o-mini & \cmark & 52.51 & 48.63 & 43.01 & 43.65 & 68.67 & 49.64 \\
GPT-4o & \cmark & \underline{63.03} & 55.76 & 47.67 & 50.45 & \textbf{81.86} & 57.97 \\
\bottomrule
\end{tabular}%
}
\label{tab:detail_vqa}
\end{table*}

For open source models, we performed experiments ino settings: with and without the original caption. The results show that providing the original caption improves performance in all tasks. Notably, the improvement is more significant for perception tasks compared to \qatype{} tasks. This is likely because perception tasks rely more heavily on understanding the image content, and the original caption provides helpful contextual information for interpreting the image.

\begin{figure*}[ht]
	\centering
	\includegraphics[width=0.99\textwidth]{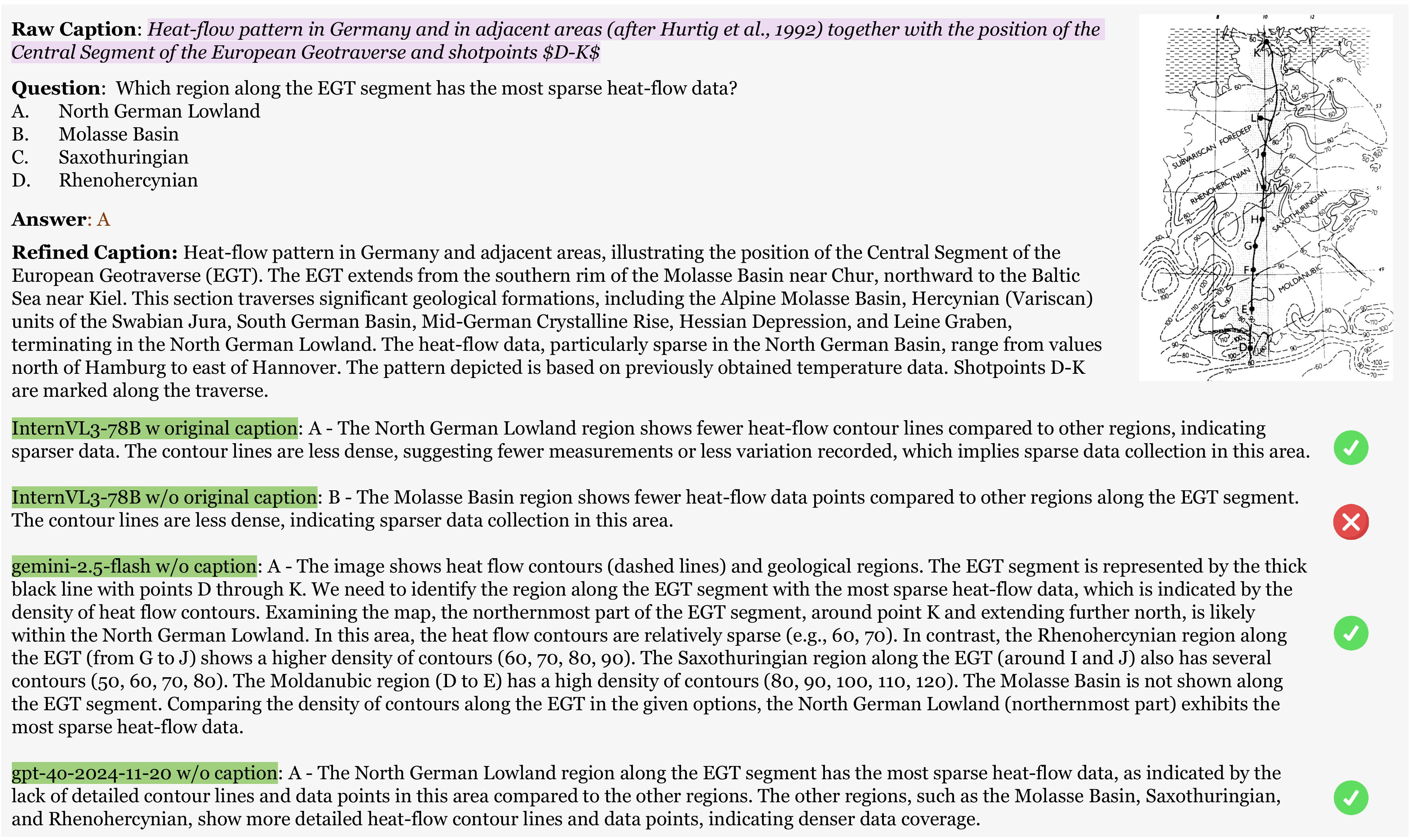}
        \caption{Performance comparison of different models under two settings: with and without the original caption. The results indicate that more powerful models exhibit less dependence on the original caption, highlighting their ability to interpret image content independently.}
        \label{case—cap1}
\end{figure*}

\begin{figure*}[ht]
	\centering
	\includegraphics[width=0.99\textwidth]{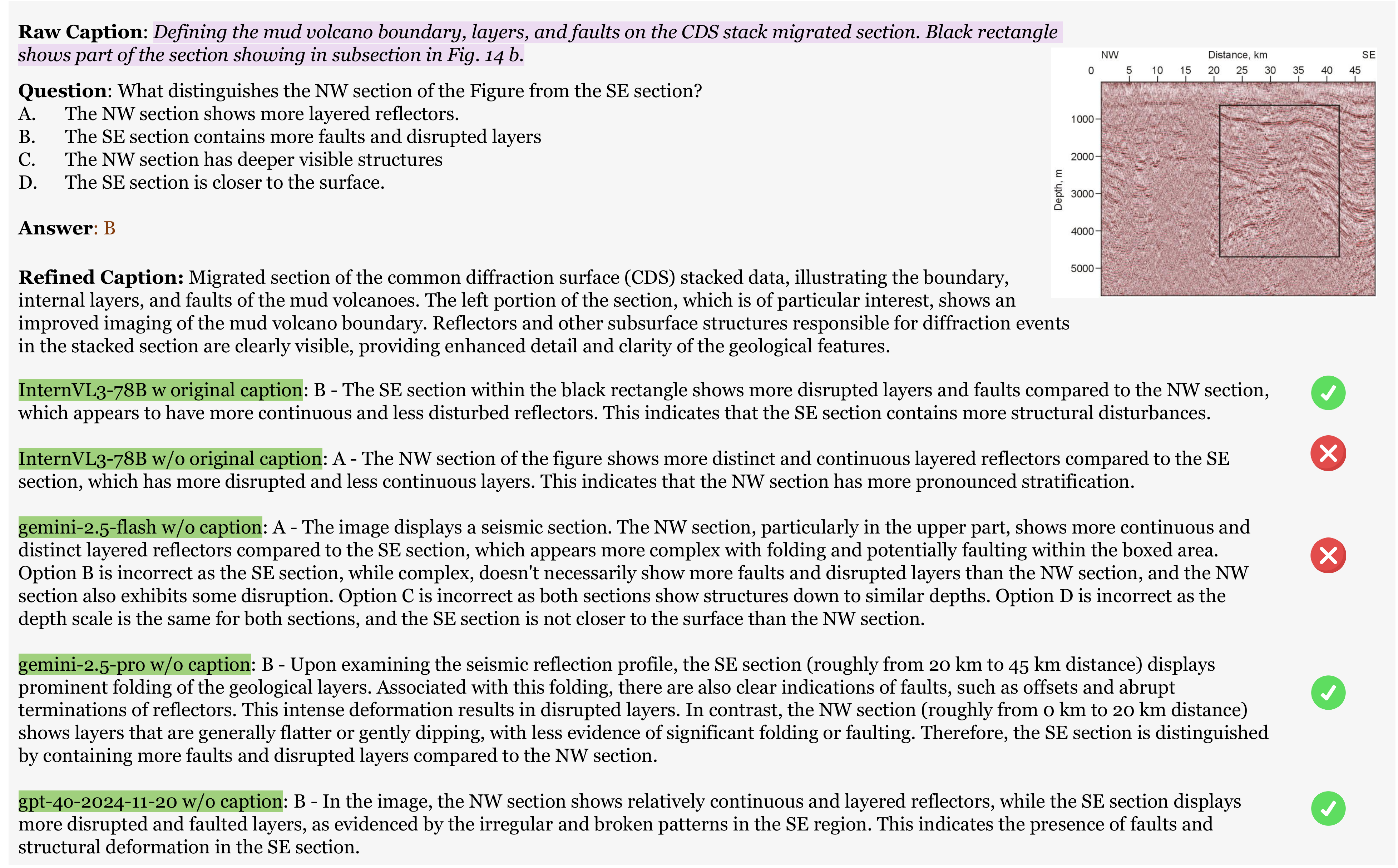}
        \caption{Performance comparison of different models under two settings: with and without the original caption. }
        \label{case—cap2}
\end{figure*}

We have compiled several case studies to illustrate the necessity of the original caption when answering questions in certain situations. In example ~\ref{case—cap1}, if the original caption is not provided, InternVL3-78B will be unable to accurately determine that the geographical location is in Germany, resulting in an incorrect answer. In contrast, some proprietary models may possess stronger perceptual capabilities and can correctly identify the location as Germany even without the original caption. Similarly, in example ~\ref{case—cap2}, providing the original caption aids the model in understanding the image, thereby facilitating task completion. Both scenarios are prevalent in scientific question-answering contexts. To address this, we conducted separate experiments and explicitly integrated these settings into the design of the MSEarth-MCQ task.

\section{Results with Compute Scaling}

\begin{figure*}[ht]
	\centering
	\includegraphics[width=\textwidth]{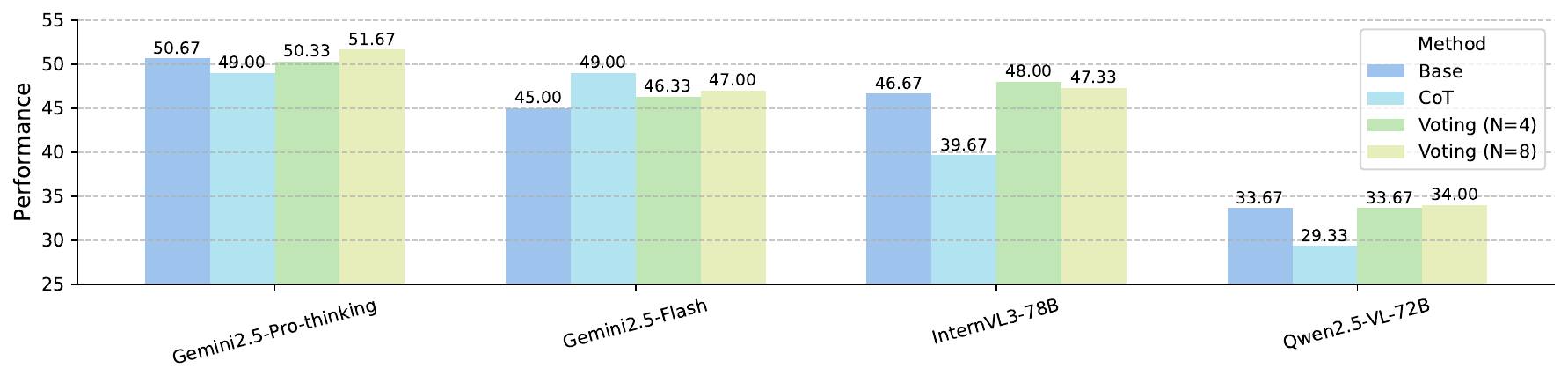}
        \caption{Performance of different strategies on \modelname{}-mini.
        }
        \vspace{-3mm}
    \label{fig: abalation}
 \end{figure*}

From the main experiments, it is evident that the performance of various models declines significantly on questions requiring specialized knowledge. To explore whether existing methods can enhance model performance on such questions, we sampled 300 specialized questions from the MCQ dataset to create the \modelname{}-mini set. We then evaluated the effectiveness of Chain-of-Thought (CoT) reasoning and majority voting mechanism, which selects the most frequent response among (N) candidate responses; in the case of a tie, one of the most frequent answers is randomly chosen. The results are presented in figure~\ref{fig: abalation}. Notably, for the Gemini-Pro-thinking model, which inherently incorporates a thinking mechanism, introducing CoT reasoning led to a decline in performance. Similarly, for some open-source models, such as Qwen and InternVL, the addition of CoT reasoning also resulted in performance degradation. However, the majority voting mechanism proved effective for most models.

\section{Impact of Explicit Reasoning}

We assess the impact of explicit chain-of-thought (CoT) prompting on MSEarth-Bench-mini across both open-source and proprietary LVLMs. For open-source models, we compare InternVL3 and QwenVL2.5; for proprietary models, we examine the Gemini-2.5-Flash series. Within the proprietary family, variants with dedicated “thinking” capabilities (e.g., Gemini-2.5-Flash-Thinking) generally outperform counterparts without such capabilities (e.g., Gemini-2.5-Flash). In contrast, for open-source models, adding explicit CoT sometimes leads to performance declines, which we hypothesize stems from limited training for explicit reasoning behaviors (e.g., GRPO-style preference optimization).

To further probe the role of explicit CoT, we include GPT-o4-mini, which exposes configuration options that control reasoning depth (low, medium, high), roughly corresponding to the length of the reasoning chain. Results are shown in Table~\ref{tab:cot_impact}.

\begin{table*}[h]
\centering
\small
\begin{tabular}{lcc}
\toprule
Model & CoT (Accuracy \%) & Non-CoT (Accuracy \%) \\
\midrule
Gemini-2.5-Pro & 50.67\% & 52.33\% \\
Gemini-2.5-Flash-no-think & 42.00\% & 40.00\% \\
Gemini-2.5-Flash-Thinking & 52.00\% & 46.00\% \\
o4-mini (low) & 52.00\% & 51.00\% \\
o4-mini (medium) & 50.67\% & 53.00\% \\
o4-mini (high) & 50.33\% & 54.33\% \\
\bottomrule
\end{tabular}
\caption{Effect of explicit chain-of-thought (CoT) prompting on MSEarth-Bench-mini. Higher is better; values are accuracy (\%). For o4-mini, low/medium/high denote shorter-to-longer reasoning traces.}
\label{tab:cot_impact}
\end{table*}

Overall, we observe the following:

Models explicitly equipped and trained for “thinking” benefit from enabling CoT (e.g., Gemini-2.5-Flash-Thinking). When a model already exhibits strong inherent reasoning, additional explicit CoT can reduce performance, as seen in o4-mini at medium/high reasoning depths. Open-source models do not consistently benefit from CoT without targeted training for reasoning behaviors, suggesting a direction for future supervised and RL post-training.

\section{More Results}
\label{add_result}

\begin{figure*}[ht]
	\centering
	\includegraphics[width=0.99\textwidth]{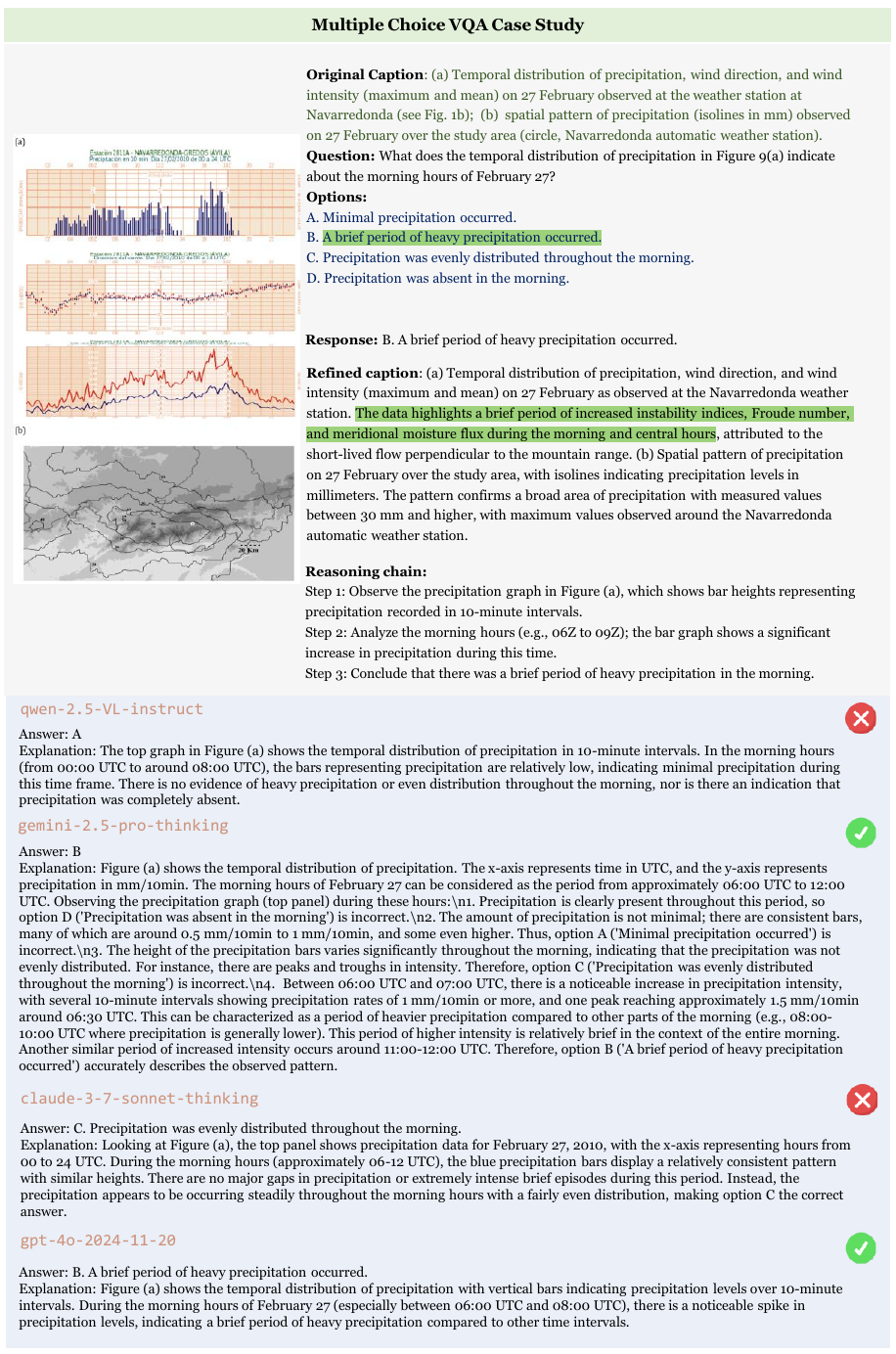}
        \caption{Case Study of Multiple Choice VQA.}
        \label{case1}
\end{figure*}

\begin{figure*}[ht]
	\centering
	\includegraphics[width=0.99\textwidth]{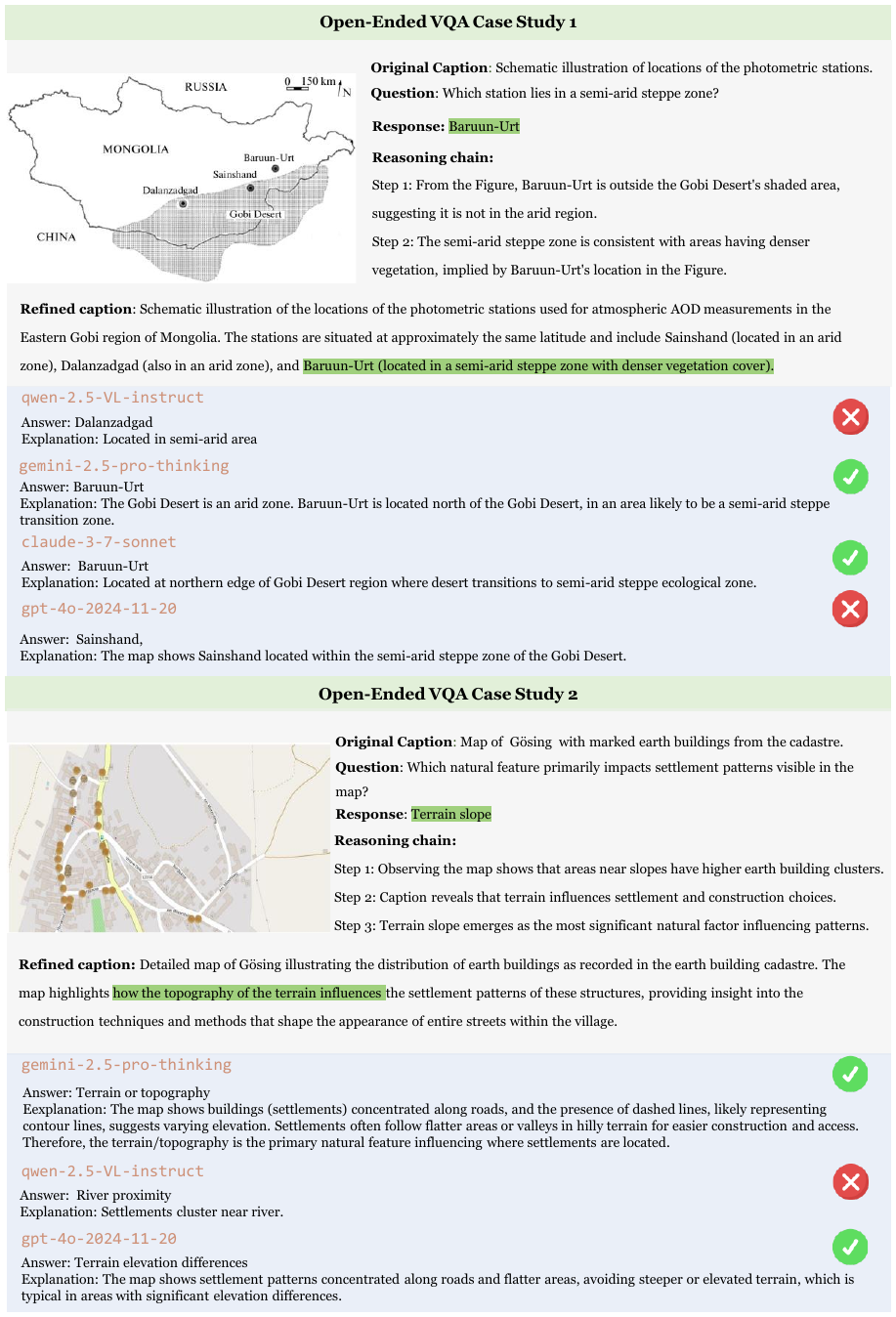}
        \caption{Case Study of Open-Ended VQA.}
        \label{case2}
\end{figure*}

For MCQ and OE questions, we used radar charts to illustrate the performance of various models across different disciplines. We also give some case studies in Figure~\ref{case1} and Figure~\ref{case2}. We also present detailed performance breakdowns of all models across every sub-discipline of Earth science in Tables~\ref{tab:subdiscipline_performance_full}.

\begin{figure*}[ht]
	\centering
	\includegraphics[width=0.99\textwidth]{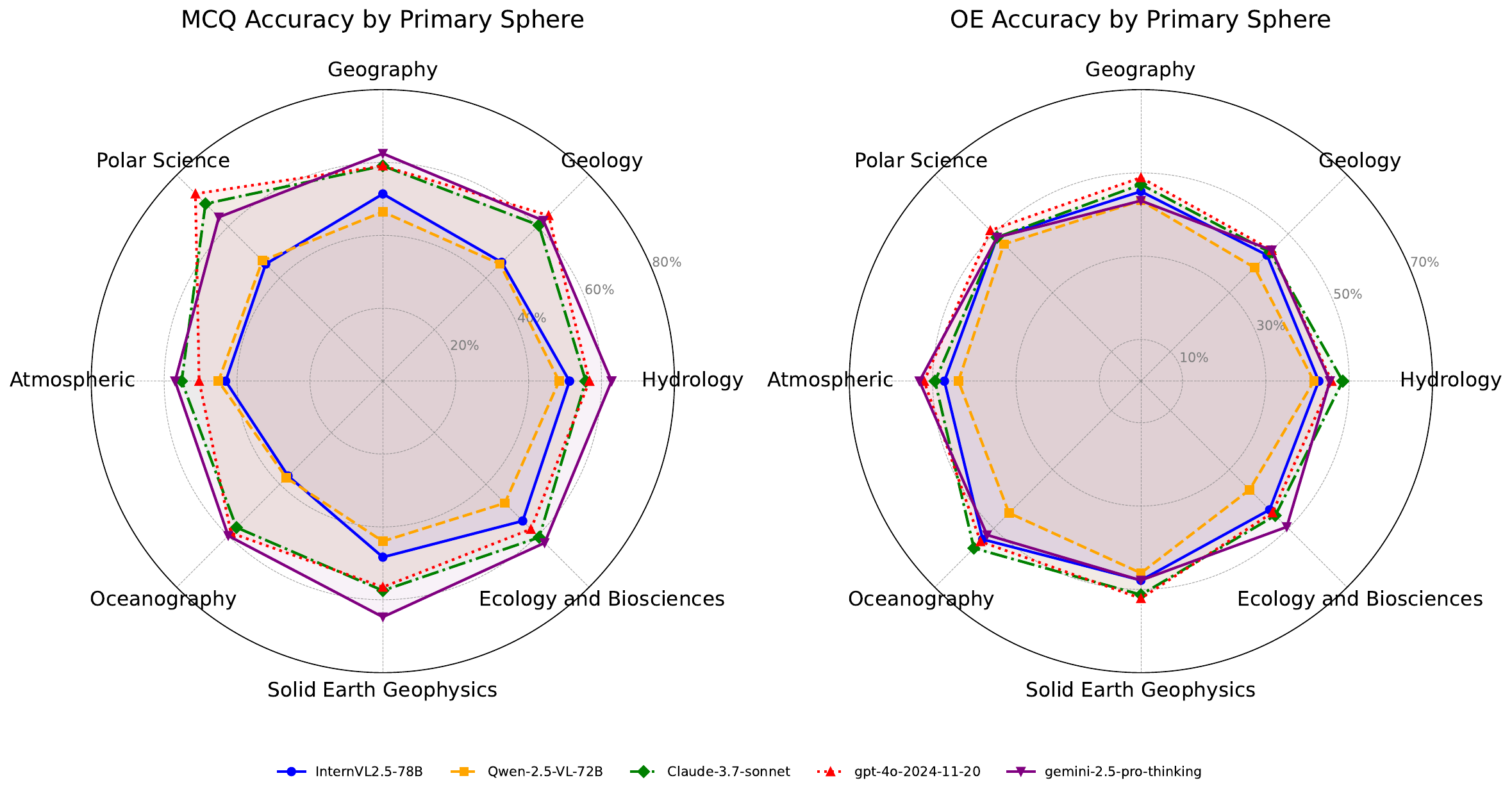}
        \caption{Performance comparison of different models across various subjects.}
        \label{sphere_case}
\end{figure*}

\begin{table*}[h]
\centering

\resizebox{\linewidth}{!}{%
\begin{tabular}{l *{6}{c}}
\hline
\textbf{Model} & 
\multicolumn{3}{c}{\textbf{Atmospheric Sciences}} & 
\multicolumn{3}{c}{\textbf{Ecology and Biosciences}} \\
\cline{2-7}
& Meteor. & Climat. & Atmos. RS & Ecosys. Ecol. & Landsc. Ecol. & Aquat. Ecol. \\
\hline
InternVL-8B & 0.4039 & 0.3643 & 0.5238 & 0.3833 & 0.4792 & 0.3030 \\
InternVL-78B & 0.4624 & 0.3643 & 0.5476 & 0.6333 & 0.5208 & 0.6061 \\
InternVL3-78B & 0.4847 & 0.3857 & 0.4762 & 0.6333 & 0.5417 & 0.6364 \\
Qwen2.5-VL-72B & 0.4903 & 0.4071 & 0.4524 & 0.4500 & 0.4375 & 0.5455 \\
Claude-3.7-Sonnet & 0.5599 & 0.4714 & 0.6429 & 0.6500 & 0.6042 & 0.5758 \\
Gemini-2.5-Pro-Thinking & 0.5877 & 0.5071 & 0.5714 & 0.7000 & 0.5833 & 0.6667 \\
GPT-4o & 0.5097 & 0.4429 & 0.5714 & 0.5667 & 0.5208 & 0.6667 \\
GPT-4o-mini & 0.4457 & 0.3857 & 0.5000 & 0.6333 & 0.4375 & 0.5455 \\
Gemini-2.5-Flash-Thinking & 0.5265 & 0.3929 & 0.5476 & 0.7500 & 0.5833 & 0.6364 \\
Gemini-2.5-Flash & 0.5153 & 0.4000 & 0.5952 & 0.6167 & 0.5625 & 0.5455 \\
Intern-S1 & 0.6320 & 0.5870 & 0.6750 & 0.7350 & 0.6580 & 0.7020 \\
Intern-S1-mini-MSEarth & 0.6080 & 0.5630 & 0.6420 & 0.7120 & 0.6350 & 0.6780 \\
Intern-S1-mini & 0.5850 & 0.5310 & 0.6180 & 0.6870 & 0.6090 & 0.6450 \\
\hline
\end{tabular}}
\caption{{Model Performance on Primary and Sub-Disciplines of Earth Science (Accuracy)}}
\end{table*}

\begin{table*}[h]
\centering
\resizebox{\linewidth}{!}{%
\begin{tabular}{l *{6}{c}}
\hline
\textbf{Model} & 
\multicolumn{3}{c}{\textbf{Geography}} & 
\multicolumn{3}{c}{\textbf{Geology}} \\
\cline{2-7}
& Phys. Geog. & Urban Geog. & Reg. Geog. & Sediment. & Struct. Geol. & Quat. Geol. \\
\hline
InternVL-8B & 0.4765 & 0.4444 & 0.4444 & 0.5090 & 0.4268 & 0.4878 \\
InternVL-78B & 0.5451 & 0.4815 & 0.5556 & 0.5663 & 0.4634 & 0.5610 \\
InternVL3-78B & 0.5884 & 0.3333 & 0.6667 & 0.5986 & 0.5366 & 0.6585 \\
Qwen2.5-VL-72B & 0.5307 & 0.3704 & 0.6667 & 0.5305 & 0.5000 & 0.6098 \\
Claude-3.7-Sonnet & 0.5993 & 0.4074 & 0.8889 & 0.6057 & 0.6341 & 0.4878 \\
Gemini-2.5-Pro-Thinking & 0.6354 & 0.5556 & 0.5556 & 0.6487 & 0.5854 & 0.6341 \\
GPT-4o & 0.5884 & 0.5556 & 0.7778 & 0.6703 & 0.6341 & 0.6829 \\
GPT-4o-mini & 0.5090 & 0.4815 & 0.6667 & 0.5269 & 0.5488 & 0.5122 \\
Gemini-2.5-Flash-Thinking & 0.6426 & 0.5556 & 0.7778 & 0.5986 & 0.5976 & 0.5122 \\
Gemini-2.5-Flash & 0.5921 & 0.5185 & 0.7778 & 0.5771 & 0.5366 & 0.5854 \\
Intern-S1 & 0.6830 & 0.6250 & 0.9120 & 0.6970 & 0.6780 & 0.7250 \\
Intern-S1-mini-MSEarth & 0.6590 & 0.5980 & 0.8850 & 0.6730 & 0.6540 & 0.6980 \\
Intern-S1-mini & 0.6320 & 0.5640 & 0.8530 & 0.6450 & 0.6210 & 0.6670 \\
\hline
\end{tabular}}
\caption{{Model Performance on Primary and Sub-Disciplines of Earth Science (Accuracy, Continued)}}
\end{table*}

\begin{table*}[h]
\centering
\resizebox{\linewidth}{!}{%
\begin{tabular}{l *{6}{c}}
\hline
\textbf{Model} & 
\multicolumn{3}{c}{\textbf{Hydrology}} & 
\multicolumn{3}{c}{\textbf{Oceanography}} \\
\cline{2-7}
& River Hydrol. & Groundw. Hydrol. & Limnol. & Ocean Phys. & Ocean Geol. & Env. Oceanogr. \\
\hline
InternVL-8B & 0.4550 & 0.4297 & 0.4348 & 0.3800 & 0.4762 & 0.2941 \\
InternVL-78B & 0.5500 & 0.4688 & 0.5652 & 0.4800 & 0.5714 & 0.3529 \\
InternVL3-78B & 0.5400 & 0.4766 & 0.5652 & 0.5250 & 0.6190 & 0.5882 \\
Qwen2.5-VL-72B & 0.5850 & 0.5078 & 0.5435 & 0.4900 & 0.6667 & 0.3529 \\
Claude-3.7-Sonnet & 0.6150 & 0.5391 & 0.5435 & 0.5650 & 0.7143 & 0.3529 \\
Gemini-2.5-Pro-Thinking & 0.6700 & 0.6172 & 0.6739 & 0.5750 & 0.6667 & 0.5882 \\
GPT-4o & 0.6200 & 0.5625 & 0.5217 & 0.5800 & 0.7619 & 0.4706 \\
GPT-4o-mini & 0.5900 & 0.4766 & 0.5435 & 0.4650 & 0.5238 & 0.4706 \\
Gemini-2.5-Flash-Thinking & 0.6250 & 0.5625 & 0.5435 & 0.5000 & 0.6667 & 0.4706 \\
Gemini-2.5-Flash & 0.5700 & 0.4922 & 0.6304 & 0.4800 & 0.5238 & 0.3529 \\
Intern-S1 & 0.7050 & 0.6680 & 0.6970 & 0.6320 & 0.7950 & 0.6230 \\
Intern-S1-mini-MSEarth & 0.6820 & 0.6430 & 0.6720 & 0.6080 & 0.7680 & 0.5950 \\
Intern-S1-mini & 0.6570 & 0.6150 & 0.6450 & 0.5830 & 0.7320 & 0.5680 \\
\hline
\end{tabular}}
\caption{{Model Performance on Primary and Sub-Disciplines of Earth Science (Accuracy, Continued)}}
\end{table*}

\begin{table*}[h]
\centering
\resizebox{\linewidth}{!}{%
\begin{tabular}{l *{6}{c}}
\hline
\textbf{Model} & 
\multicolumn{3}{c}{\textbf{Polar Science}} & 
\multicolumn{3}{c}{\textbf{Solid Earth Geophysics}} \\
\cline{2-7}
& Glaciol. & Permafrost Sci. & Polar Ocean & Seismol. & Tectonophys. & Geomagn. \\
\hline
InternVL2.5-8B & 0.4571 & 0.7500 & 0.0000 & 0.4248 & 0.5625 & 0.5455 \\
InternVL2.5-78B & 0.4571 & 0.5000 & 1.0000 & 0.4902 & 0.4375 & 0.5455 \\
InternVL3-78B & 0.5286 & 0.5000 & 1.0000 & 0.5033 & 0.5000 & 0.6364 \\
Qwen2.5-VL-72B & 0.6286 & 0.5000 & 1.0000 & 0.4706 & 0.3750 & 0.7273 \\
Claude-3.7-Sonnet & 0.7000 & 0.5000 & 1.0000 & 0.5752 & 0.5625 & 0.6364 \\
Gemini-2.5-Pro-Thinking & 0.6286 & 1.0000 & 1.0000 & 0.6863 & 0.5000 & 0.8182 \\
GPT-4o & 0.7286 & 0.7500 & 1.0000 & 0.5163 & 0.6250 & 0.9091 \\
GPT-4o-mini & 0.5571 & 0.7500 & 1.0000 & 0.4379 & 0.4375 & 0.6364 \\
Gemini-2.5-Flash-Thinking & 0.6571 & 0.7500 & 1.0000 & 0.5359 & 0.6250 & 0.8182 \\
Gemini-2.5-Flash & 0.6812 & 0.7500 & 1.0000 & 0.6013 & 0.5000 & 0.5455 \\
Intern-S1 & 0.7650 & 1.0000 & 1.0000 & 0.7230 & 0.6850 & 0.9320 \\
Intern-S1-mini-MSEarth & 0.7380 & 0.9500 & 1.0000 & 0.6970 & 0.6580 & 0.9050 \\
Intern-S1-mini & 0.7120 & 0.9000 & 1.0000 & 0.6650 & 0.6230 & 0.8780 \\
\hline

\multicolumn{7}{l}{\small \textit{Note:} 1. Sub-disciplines listed are the top 3 with the largest sample size in each primary discipline;} \\
\multicolumn{7}{l}{\small 2. Abbreviations: Meteor.=Meteorology, Climat.=Climatology, Atmos. RS=Atmospheric Remote Sensing,} \\
\multicolumn{7}{l}{\small Ecosys. Ecol.=Ecosystem Ecology, Landsc. Ecol.=Landscape Ecology, Aquat. Ecol.=Aquatic \& Limnological Ecology,} \\
\multicolumn{7}{l}{\small Phys. Geog.=Physical Geography, Reg. Geog.=Regional Geography, Sediment.=Sedimentology,} \\
\multicolumn{7}{l}{\small Struct. Geol.=Structural Geology, Quat. Geol.=Quaternary Geology, River Hydrol.=River \& Estuarine Hydrology,} \\
\multicolumn{7}{l}{\small Groundw. Hydrol.=Groundwater Hydrology, Limnol.=Limnology, Env. Oceanogr.=Environmental Oceanography,} \\
\multicolumn{7}{l}{\small Glaciol.=Glaciology, Seismol.=Seismology, Tectonophys.=Tectonophysics, Geomagn.=Geomagnetism.} \\
\end{tabular}}
\caption{{Model Performance on Primary and Sub-Disciplines of Earth Science (Accuracy, Continued)}}
\label{tab:subdiscipline_performance_full}
\end{table*}

\clearpage
\newpage

\end{document}